\documentclass[10pt,twocolumn,letterpaper]{article}

\usepackage[pagenumbers]{cvpr} 

\definecolor{cvprblue}{rgb}{0.21,0.49,0.74}
\usepackage[pagebackref,breaklinks,colorlinks,allcolors=cvprblue]{hyperref}
\usepackage{amssymb}
\usepackage{pifont}
\usepackage{tabularx}
\usepackage{multirow}
\usepackage{makecell}
\usepackage{adjustbox}
\usepackage{subcaption}

\newcommand{\cmark}{\ding{51}}%
\usepackage[normalem]{ulem}

\def\paperID{} 
\def\confName{CVPR}
\def\confYear{2026}

\title{RAMEN: Resolution-Adjustable Multimodal Encoder for Earth Observation}

\author{
Nicolas Houdré \textsuperscript{1} 
\and
Diego Marcos \textsuperscript{2}
\and
Hugo Riffaud de Turckheim \textsuperscript{2}
\and
Dino Ienco \textsuperscript{3}
\and
Laurent Wendling \textsuperscript{1}
\and
Camille Kurtz \textsuperscript{1}
\and
Sylvain Lobry \textsuperscript{1}
}
\begin{document}
\twocolumn[
  \begin{@twocolumnfalse}
    \maketitle
    \vspace{-3em}
    \begin{center}
        \textsuperscript{1} Université Paris Cité, LIPADE, F-75006 Paris, France;  
        \textsuperscript{2} UMR TETIS, EVERGREEN, Inria, Univ. Montpellier, Montpellier, France;
        \textsuperscript{3} UMR TETIS, EVERGREEN, INRAE, Inria, Univ. Montpellier, Montpellier, France
    
    \end{center}
  \end{@twocolumnfalse}
  \vspace{1em} 
]

\begin{abstract}
Earth observation (EO) data spans a wide range of spatial, spectral, and temporal resolutions, from high-resolution optical imagery to low resolution multispectral products or radar time series. 
While recent foundation models have improved multimodal integration for learning meaningful representations, they often expect fixed input resolutions or are based on sensor-specific encoders limiting generalization across heterogeneous EO modalities.
To overcome these limitations we introduce RAMEN, a resolution-adjustable multimodal encoder that learns a shared visual representation across EO data in a fully sensor-agnostic manner. 
RAMEN treats the modality and spatial and temporal resolutions as key input data features, enabling coherent analysis across modalities within a unified latent space. 
Its main methodological contribution is to define spatial resolution as a controllable output parameter, giving users direct control over the desired level of detail at inference and allowing explicit trade-offs between spatial precision and computational cost. 
We train a single, unified transformer encoder reconstructing masked multimodal EO data drawn from diverse sources, ensuring generalization across sensors and resolutions.
Once pretrained, RAMEN transfers effectively to both known and unseen sensor configurations and outperforms larger state-of-the-art models on the community-standard PANGAEA benchmark, containing various multi-sensor and multi-resolution downstream tasks.
Our code and pretrained model are available at \url{https://github.com/nicolashoudre/RAMEN}.
\end{abstract}    
\section{Introduction}
\label{sec:intro}

\begin{figure}[!t]
    \centering
    \includegraphics[width=\linewidth]{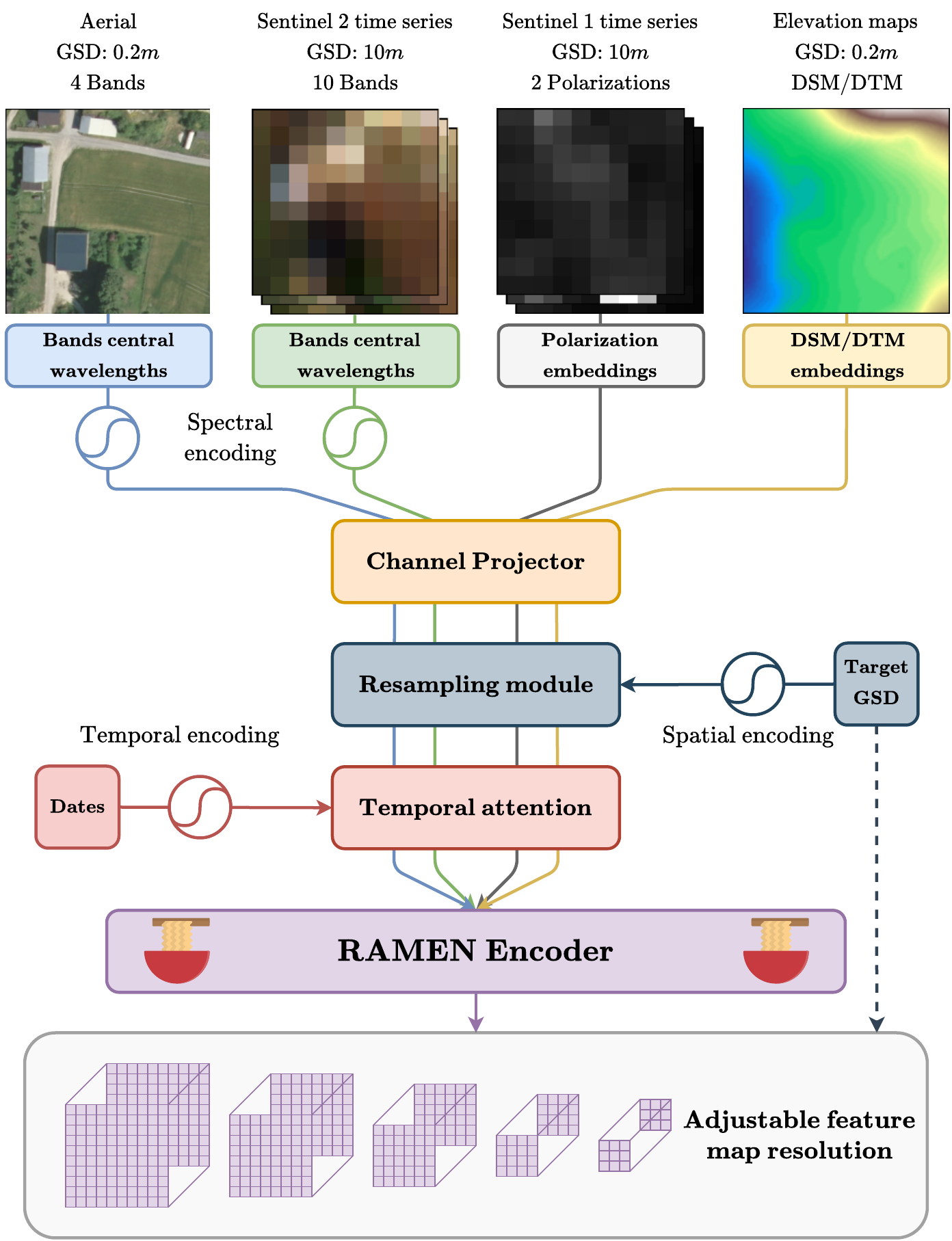}
    \caption{\textbf{Visual workflow of RAMEN.}
    RAMEN enables consistent adaptation across multimodal imagery via resolution-specific projection modules. Thanks to our \textbf{adjustable resampling} strategy and pretraining scheme, RAMEN allows practitioners to define the feature map spatial resolution of encoded inputs, allowing fine-grained representations and trade-offs between downstream performances and computational overhead.}
    \label{fig:intro_ramen}
\end{figure}

Earth observation (EO) data is inherently heterogeneous. Remote sensing (RS) data ranging from aerial RGB cameras to multispectral satellites, SAR sensors, or elevation maps exhibit significant variability in terms of channel characteristics, ground sampling distance (GSD - distance between pixel centers measured on the ground), and temporal sampling patterns. This diversity, while challenging to construct a unified model, is what enables EO data to serve a wide range of applications. Tasks such as land-cover mapping~\cite{garnot2021panoptic, m2019semantic, toker2022dynamicearthnet}, disaster response~\cite{rambour2020flood, jakubik2023foundation} or urban planning~\cite{van2018spacenet} rely on different combinations of spatial detail, spectral information or temporal dynamics. 
Leveraging this large volume of heterogeneous data in a coherent manner remains a core challenge for general-purpose EO models.
Several multimodal EO foundation models (FMs) tackled these issues by integrating sensor-specific encoders \citep{CROMA, SatMAE, seco}.
Using these models on new modalities requires changing and retraining parts of the architecture, thus limiting the generalization capabilities of such models. Recent efforts on EO FMs enabled more flexibility in terms of pretraining data~\citep{satlas}, multimodal integration~\citep{AnySat, Galileo}, or spatial~\citep{Scale-MAE, FlexiMo} and spectral~\cite{SMARTIES, DOFA} resolution handling.
However, these models still fall short of fully accounting for the diverse characteristics of EO sensors. They often neglect differences in modal, spatial and temporal properties and produce feature representations at fixed resolutions, limiting their scalability across tasks requiring different spatial detail or computational efficiency.

To address these challenges, we propose a \textbf{R}esolution-\textbf{A}djustable \textbf{M}ultimodal \textbf{En}coder (\textbf{RAMEN})  that treats every spatio-spectro-temporal resolution aspect of modalities and project representations in a resolution-aware shared space. As illustrated in Fig.~\ref{fig:intro_ramen}, this enables RAMEN to learn a unified transformer encoder on a wide combination of heterogeneous modalities from diverse datasets. Each modality, with its own channels and spatial and time resolutions, is treated separately to produce a universal multimodal representation: 
(i) A channel-conditioned projector based on each band's central wavelength (or radar polarization/elevation maps) projects every input to the latent space dimension; 
(ii) An adjustable spatial resampler module enables alignment of spatially heterogeneous inputs by mapping them to a target GSD. 
At pretraining time, this target GSD is chosen randomly for each modality and dataset, ensuring that the encoder learns to generalize across scales and is not biased toward a specific setting. 
Such a mechanism further allows practitioners to explicitly choose the desired resolution during fine-tuning or inference depending on the downstream task and computational resources available; 
(iii) A temporal attention module enriched with temporal positional encoding captures dependencies across time. 
Together, these components allow RAMEN to optimize multimodal representations that are robust to heterogeneous spatio-spectro-temporal characteristics.
Leveraging these unified multimodal representations, we pretrain RAMEN using a large-scale multimodal EO corpus combining multiple datasets in a self-supervised manner using a masked image modeling~\cite{mae} strategy by reconstructing randomly masked regions of each modality representation in the resolution-aware space.
Besides type-aware embeddings covering three major input types (i.e. spectral imagery, radar imagery and elevation maps), all learnable parameters in our model are shared across all modalities, enabling an architecture trainable on highly heterogeneous multimodal EO datasets. 
Finally, RAMEN also provides users with explicit control over desired spatial resolution of the representations at inference, allowing them to balance task-specific needs with available computational resources and desired precision. 
We evaluate RAMEN on eight datasets from the community-standard PANGAEA~\cite{PANGAEA} benchmark, covering diverse uni- and multimodal semantic segmentation tasks, and show that it delivers state-of-the-art performance, strong flexibility and robust generalization across resolutions - including those unseen during pretraining.
Our contributions can be summarized as follows:
\begin{itemize}
    \item We propose RAMEN, a resolution-adjustable encoder that can process remote sensing images of any sensors and configurations without any retraining;
    
     \item The proposed architecture allows practitioners to configure the desired spatial resolution at inference time, enabling flexible adaptation of the compute/performance trade-off to the need of the task;
    
    \item We demonstrate that when pretrained on a large collection of diverse modalities, RAMEN outperforms the state-of-the-art on standard EO benchmark datasets.
    
\end{itemize}
\section{Related work}
\label{sec:relatedwork}

In this section, we highlight recent supervision and architecture advancements designed to support multimodal FMs in remote sensing, particularly dealing with multi-resolution setup and their adaptability to diverse input data. 

\noindent\textbf{Self-supervised learning (SSL)} has emerged as a powerful paradigm to learn generalizable representations from large non-annotated datasets transferable to various downstream tasks. Among SSL approaches, masked image modelling~\citep{mae} aims at reconstructing randomly masked regions of an input, encouraging the model to learn rich semantic structures without requiring specific data augmentation.
The abundance of raw EO data has naturally led to the adaptation of such methods for RS. In this context, \emph{monomodal SSL} approaches pretrain models on single sensor data, often optical imagery~\cite{SatMAE, SatMAEpp, RingMo, Prithvi, SpectralGPT}. While effective to learn meaningful representations from large-scale EO datasets, these methods remain inherently constrained to specific assumptions on input data. 
To move beyond this limitation, \emph{multimodal SSL} aims at jointly leveraging heterogeneous EO sensors (e.g. multispectral and radar imagery) through sensor-specific encoders~\cite{CROMA, OmniSat, DeCUR} or projections in a unified representation space~\cite{MMEarth, Presto, EarthView, GFM, TerraMind, SkySense, MAESTRO}.
However, many FMs rely on common computer vision architectures not explicitly designed for the specificities of EO data such as heterogeneous modalities (e.g. multispectral, radar, elevation maps), variable resolutions or spatio-temporal dependencies~\cite{AIEOreview}.

\noindent\textbf{Spectrally-aware FMs.}
While most EO foundation models process multispectral imagery as generic multi-channel inputs, recent works have explored incorporating the physical meaning of spectral bands into model design.
DOFA~\cite{DOFA} authors introduce a wavelength-conditioned hypernetwork that dynamically generates convolutional weights based on the central wavelength of each input band.
Similarly, SMARTIES~\cite{SMARTIES} or Panopticon~\cite{Panopticon} use the central wavelength of each band to respectively guide spectrum-aware projection layers or attention over channels.

\noindent\textbf{Spatially-aware FMs.}
Several architectures have been proposed to address the varying spatial resolution of EO images. Notably, Scale-MAE~\cite{Scale-MAE} introduced GSD-based positional encodings, allowing the model to adapt feature extraction based on input resolution.
AnySat~\cite{AnySat} maintains spatial awareness by considering patch sizes for each modality representing real ground distances rather than pixels, but introduce modality-specific projectors reducing generalization capabilities of the model. 
FlexiMo~\cite{FlexiMo} builds on the theory introduced by~\cite{FlexiViT} to dynamically adapt both raw images and patch embedding convolution weights.

\noindent\textbf{Temporally-aware FMs.}
Temporal information plays a central role in remote sensing, as many EO applications such as crop monitoring or disaster response rely on multi-date observations. This aspect has motivated several EO foundation models to explicitly model temporal dynamics.
PRESTO~\cite{Presto} and Galileo~\cite{Galileo} process each time-steps as separate token sequences, while Prithvi~\cite{Prithvi} treats spatio-temporal sequences using 3D patch embeddings coupled with 3D positional embeddings for time, height and width.
AnySat employs a Lightweight Temporal Attention Encoder (LTAE)~\cite{LTAE} to embed patches coming from temporal modalities. 

\noindent\textbf{Generalization.}
Existing multimodal EO foundation models typically handle only specific aspects of resolution heterogeneity: some focus on spectral variability (e.g. DOFA, SMARTIES), while others incorporate spatial awareness (e.g. Scale-MAE, FlexiMo) or temporal dynamics (e.g., AnySat, Galileo). 
Yet none address all three aspects jointly. Compared to previous works, RAMEN is the first model to provide a modality-agnostic, multi-temporal and resolution-adjustable framework.
\section{RAMEN: Resolution-Adjustable Multimodal Encoder}

In this section, we introduce the RAMEN architecture (see also Fig.~\ref{fig:architecture}). We first describe the resolution aware modules treating the channels of the different modalities, as well as spatial and temporal resolutions and their integration in a multimodal network in Sec.~\ref{sec:architecture}. 
Then, we detail our self-supervised pretraining scheme {in }Sec.~\ref{sec:ssl-training}.

\subsection{Architecture}\label{sec:architecture}
A central challenge in RS is that different sensors exhibit differences along three axes: \emph{modality} (e.g. RGB or multispectral optical bands, radar polarizations or digital elevation maps), 
\emph{spatial} (e.g. 0.2\,m aerial imagery vs. 10-30\,m satellite acquisitions) and \emph{temporal} (e.g. sparsely acquired aerial imagery or regular satellite revisits). Therefore, direct multimodal alignment or concatenation is often infeasible due to the heterogeneous dimensions.

To unify such heterogeneous data, RAMEN introduces three modules encoding each of these input axes (modality, spatial and temporal) in a shared latent space while preserving their physical meaning. 

Formally, we consider a set of geospatially aligned images $(x_1, \dots, x_M)$ observed through $M$ modalities. Each input image consists of a tensor $x_m \in \mathbb{R}^{T_m \times C_m \times H_m \times W_m}$ where $m$, $T_m$, $C_m$, $H_m$, $W_m$ represent the considered modality, number of temporal observations, number of channels, height and width, respectively.

\begin{figure*}[!t]
    \centering
    \includegraphics[width=\textwidth]{figures/MAE.pdf}
    \caption{\textbf{Architecture of RAMEN.}
    At each iteration, a subset of modalities and a target ground sampling distance (GSD) are sampled. Each selected modality is projected into a shared latent space through three resolution-specific modules: 
    (a) A \textbf{channel-conditioned projector} that embeds the physical meaning of each channel;
    (b) An \textbf{adjustable spatial resampler} maps features to the user-defined $GSD_{target}$ with scale-aware convolutional adaptation; 
    (c) A \textbf{temporal attention module} treats time-series enriched with the day of acquisition encoding.
    We adopt a masked image modeling pretraining scheme, where we reconstruct each modality at its native spectral, spatial and temporal resolution with inverted modules.
    }
    \label{fig:architecture}
\end{figure*}

\noindent\textbf{Channel-conditioned projector.}\label{sec:spectralproj}
EO sensors differ, not only in the number of channels captured, but also in their physical interpretation. To unify these heterogeneous channel configurations, RAMEN adopts channel-wise embeddings carrying each band's physical meaning.
For optical and multispectral modalities, we follow the DOFA~\cite{DOFA} formulation and embed the central wavelength of each channel through a sinusoidal positional encoding~\cite{transformer}:
\begin{align}
    \text{PE}(\lambda^i_m, 2k) &= \sin\frac{\lambda^i_m}{10000^{2k/D}}\,; \\
    \text{PE}(\lambda^i_m, 2k+1) &= \cos\frac{\lambda^i_m}{10000^{2k/D}}\,,
\end{align}
where $\lambda^i_m$ is the central wavelength of channel $i$ in nanometers and $D$ the latent space dimension.

For non-optical modalities, we use a specific learned embedding, also of size $D$, corresponding to each of the available channels in the considered datasets: the four transmit/receive polarizations for SAR images (VV, VH, HH and HV) in both ascending and descending orbits, as well as the digital surface (DSM), terrain (DTM) and slope models for elevation data.

These encodings are concatenated and processed by a lightweight multi-layer perceptron (MLP) to obtain a channel-wise projection matrix $M_m \in \mathbb{R}^{C_m \times D}$ that maps raw inputs to the latent space as:
\begin{equation}
    x^S_m(t,d,h,w) = \sum^{C_m}_{c=1} x_m(t,c,h,w)M_m(c,d)\,,
\end{equation}
where $x^S_m \in \mathbb{R}^{T_m \times D \times H_m \times W_m}$.

\noindent\textbf{Spatial resampler.}
To unify features originating from sensors with different native spatial resolutions, we design an \emph{adjustable spatial resampler} that maps each spectrally projected features $x^S_m \in \mathbb{R}^{T_m \times D \times H_m \times W_m}$ (acquired at a ground sampling distance $\text{GSD}_{m}$) to a user-defined target resolution $\text{GSD}_{target}$. The $\text{GSD}_{target}$ can be defined at inference time based on the downstream task of interest.
Because sensors can differ by orders of magnitude in spatial resolution, we introduce a \emph{mixture of $N_{conv}$ convolutions} mechanism that dynamically adapts interpolated images based on the interpolation ratio between the input and target resolutions. 

Let $\sigma_{m} = \log (\text{GSD}_{m}/\text{GSD}_{target})$ denote the log-scaled interpolation ratio. This scalar quantifies how much the spatial resolution of modality $m$ differs from the target resolution in the log-space, ensuring both down/upsampling are treated symmetrically. 
Each convolution operator, or expert, 
is implemented as a single $1\times1$ convolution. Conditioning their mixture on $\sigma_m$ enables the model to select how to refine interpolated features depending on the magnitude and direction of the scale change. 
This provides a lightweight yet expressive mechanism to correct feature statistics after interpolation without altering spatial structure.

Given $\mathcal{I}_{\sigma_{m}}$ a bilinear interpolation operator parameterized by $\sigma_{m}$, and $\text{Conv}_{n}$ the $n^{th}$ expert convolution, the final spatially aligned representation $x^R_m \in \mathbb{R}^{T_m \times D \times H_{target} \times W_{target}}$ is defined as:
\begin{equation}
    x^R_m = \mathcal{I}_{\sigma_{m}}(x^S_m) + \sum_{n=1}^{N_{\text{conv}}} {w}_n {\text{Conv}}_{n}( \mathcal{I}_{\sigma_{m}}(x^S_m))\,,
\end{equation}
where $w_n$ is the weight assigned to the $n^{th}$ expert, $H_{target} = \exp(\sigma_m)H_m$ and $W_{target} = \exp(\sigma_m)W_m$.

Analogously to spectral encodings, we embed $\sigma_{m}$ into a  $D$-dimensional vector using sinusoidal positional encoding.
The resulting embedding is passed through a MLP followed by a softmax layer to produce normalized weights:
\begin{equation}
    ({w}_1, \dots, {w}_{N_{conv}}), \text{with} \sum_{n=1}^{N_{conv}} {w}_n = 1\,.
\end{equation}

This formulation enables continuous interpolation across diverse spatial resolutions while refining spatial details through scale-aware convolutional adaptation.

\noindent\textbf{Temporal attention.}
To process time-series and model temporal dependencies, we employ a Lightweight Temporal Attention Encoder (LTAE)~\cite{LTAE}. 
To preserve temporal continuity, a sine-cosine positional encoding based on the day of acquisition is added to each timestamp. Self-attention along the temporal axis is then applied to spectrally and spatially projected features $x^R_m$, resulting in a temporally aggregated representation:
\begin{equation}
    x^T_m = \text{LTAE}(x^R_m), \text{with~~} x^T_m \in \mathbb{R}^{D \times H_{target} \times W_{target}}\,.
\end{equation}

\noindent\textbf{Multimodal network.}
After temporal aggregation, we obtain a set of feature maps $\{x^T_m\}_{m=1}^{M}$, each representing a different modality projected into the shared resolution-aware latent space.
Each spatial position is treated as an individual token, such that one pixel at ground sampling distance $\text{GSD}_{target}$ corresponds to one $D$-dimensional embedding.

Similarly to Scale-MAE \cite{Scale-MAE}, we enrich these tokens with GSD-based positional encodings that convey information about the target spatial resolution $\text{GSD}_{target}$. We note that this approach combined with our adjustable spatial resampler and interpolation ratio encoding provides implicit knowledge about the original input spatial resolution as well. 

All modality-specific tokens are then flattened along the sequence dimension and concatenated to produce the final multimodal sequence $Z \in \mathbb{R}^{N \times D}$, where $N =  M \cdot H_{\text{target}} \cdot W_{\text{target}}$ 
denotes the total number of tokens across modalities.
These multimodal tokens are jointly processed by a shared Transformer~\cite{transformer} encoder, enabling cross-modal reasoning and global spatio-spectro-temporal interactions without requiring modality-specific branches.

\subsection{Self-supervised training}\label{sec:ssl-training}

RAMEN is trained using a masked image modeling through masked autoencoders (MAE) objective \cite{mae} applied to the unified multimodal token sequence $Z$.
The goal is to reconstruct all modalities at their native spatial, channel, and temporal resolutions, using only a subset of visible tokens in the latent space. This forces the model to learn resolution-consistent and modality-agnostic representations.

\noindent\textbf{Encoding and decoding.}
We apply random masking with ratio $R$ uniformly across all tokens in the multimodal sequence $Z$.
Visible tokens and a learnable $[\text{CLS}]$ token are processed by a ViT encoder in the unified resolution-aware latent space. 
Following MAE formulation, the decoder receives the full set of tokens composed of: (i) the encoded visible tokens, (ii) shared mask tokens leveraged by positional encodings.

\noindent\textbf{Resolution-adjustable image reconstruction.}
The obtained decoded representations are projected back to their modality-specific resolutions via dedicated \emph{reconstruction} modules that mirror their corresponding projection components introduced in Sec.~\ref{sec:architecture}.
Reconstruction proceeds in three steps:
(i) Temporal expansion: for each modality $m$, its sequence is expanded back to the temporal dimension $T_m$. Each time step is enriched with day of acquisition encoding and fed into a single transformer block;
(ii) Spatial resampling: features are mapped from their $\text{GSD}_{target}$ spatial resolution back to $\text{GSD}_m$ with a spatial resampler module;
(iii) Channel reconstruction: modality-specific channels are recovered from the embedding dimension via the transpose of the channel-wise projection matrix produced by the channel-conditioned projector.

We supervise the model using the Mean Squared Error (MSE) loss between the original masked pixels $x^{masked}_m$ and their reconstructed equivalent $\hat{x}^{masked}_m$, yielding the final reconstruction loss:

\begin{equation}
    \mathcal{L} = \frac{1}{M} \sum_{m=1}^M \frac{\left(\hat{x}^{masked}_m - x^{masked}_m\right)^2}{H_m W_m}\,.
\end{equation}

\noindent\textbf{Generalizing on heterogeneous modalities across datasets.}
RAMEN is designed to flexibly encode arbitrary combinations of multimodal inputs regardless of their spatio-spectro-temporal characteristics. To fully leverage this resolution adjustable encoding during pretraining, we employ a \emph{random multimodal sampling strategy} on a large heterogeneous corpus of multimodal EO datasets.

At each pretraining iteration:
\begin{enumerate}
    \item a dataset is randomly sampled from the multimodal corpus;  
    \item a subset of modalities available in that dataset is randomly selected;
    \item a target $\text{GSD}_{target}$ is drawn from a dataset-specific range of resolutions;  
    \item all sampled modalities are projected into the shared latent space using the modules described in Sec.~\ref{sec:architecture}.
\end{enumerate}

This stochastic strategy has two desirable effects. First, it exposes RAMEN to a continuously varying combination of sensors and spatio-spectro-temporal resolutions, encouraging the model to learn modality-agnostic and resolution-consistent representations. 
Second, it substantially improves computational efficiency: high resolution multimodal sequences that are the most memory- and time-demanding are only sampled occasionally, while the model learns to adapt and generalize to the full spectrum of available resolutions and combinations.

\section{Experiments}

\begin{table*}[!t]
\centering
\caption{\textbf{PANGAEA \cite{PANGAEA} benchmark results on 8 EO downstream tasks}. 
The best model per column is highlighted in bold, the second best is underscored. We indicate unimodal datasets with $^\ast$. Encoders are frozen for pretrained models, while U-Net and ViT baselines are trained from scratch for each specific task.}
\begin{adjustbox}{width=1\textwidth}
\begin{tabular}{lccccccccc|cc}
\toprule
Model & Model size & BurnSr$^\ast$~\cite{jakubik2023foundation} & MADOS$^\ast$~\cite{kikaki2024detecting} & PASTIS~\cite{garnot2021panoptic} & Sen1Fl11~\cite{rambour2020flood} & DEN$^\ast$~\cite{toker2022dynamicearthnet} & CTM-SS~\cite{m2019semantic} & SN7$^\ast$~\cite{van2018spacenet} & AI4Farms$^\ast$~\cite{ai4small} & Avg. mIoU & Avg. Rank \\
\midrule
U-Net baseline~\cite{ronneberger2015u} & - &  \underline{84.51} & 54.79  & 31.60 & \textbf{91.42}  & \underline{39.46} & 47.57 & \underline{62.09} & \textbf{46.34} & 57.22 & 4.25 \\
ViT baseline~\cite{vit} & Base       & 81.58        & 48.19  & 38.53 & 87.66  & 36.83 & 44.08 & 52.57 & 38.37 & 53.48 & 10.88 \\
\midrule
CROMA~\cite{CROMA} & Large            & 82.42        & 67.55  & 32.32 & 90.89 & 38.29 & 49.38 & 59.28 & 25.65 & 55.72 & 6.50 \\
DOFA~\cite{DOFA} & Base                & 80.63        & 59.58  & 30.02 & 89.37 & 39.29 & 51.33 & 61.84 & 27.07 & 54.89 & 7.50 \\
GFM-Swin~\cite{GFM} & Base            & 76.90        & 64.71  & 21.24 & 72.60  & 34.09 & 46.98 & 60.89 & 27.19 & 50.57 & 11.00 \\
Prithvi 1.0 100M~\cite{Prithvi} & Base   & 83.62       & 49.98  & 33.93 & 90.37 & 27.86 & 43.07 & 56.54 & 26.86 & 51.53 & 10.88 \\
RemoteCLIP~\cite{RemoteCLIP} & Base         & 76.59        & 60.00  & 18.23 & 74.26 & 31.78 & 52.05 & 57.76 & 25.12 & 49.47 & 12.63 \\
SatlasNet~\cite{satlas} & Base           & 79.96        & 55.86  & 17.51 & 90.30 & 36.31 & 46.97 & 61.88 & 25.13 & 51.74 & 10.63 \\
Scale-MAE~\cite{Scale-MAE} & Large          & 76.68        & 57.32  & 24.55 & 74.13 & 35.11 & 25.42 & \textbf{62.96} & 21.47 & 47.21 & 12.75 \\
SpectralGPT~\cite{SpectralGPT} & Base        & 80.47        & 57.99  & 35.44 & 89.07 & 37.85 & 46.95 & 58.86 & 26.75 & 54.17 & 9.50 \\
S.-S12-MoCo~\cite{moco} & Small       & 81.58        & 51.76  & 34.49 & 89.26 & 35.44 & 48.58 & 57.64 & 25.38 & 53.02 & 10.50 \\
S.-S12-DINO~\cite{dino} & Small       & 81.72        & 49.37  & 36.18 & 88.61 & 34.81 & 48.66 & 56.47 & 25.62 & 52.68 & 11.00 \\
S.-S12-MAE~\cite{mae} & Small        & 81.91        & 49.90  & 32.03 & 87.79 & 34.08 & 45.80 & 57.13 & 24.69 & 51.67 & 12.88 \\
S.-S12-Data2Vec~\cite{data2vec} & Small   & 81.91        & 44.36  & 34.32 & 88.15 & 35.90 & 54.03 & 58.23 & 24.23 & 52.64 & 10.63 \\
TerraMindv1-B~\cite{TerraMind} & Base     & 82.42        & 69.52 & 40.51 & 90.62 & 37.87 & \textbf{55.80} & 60.61 & 28.12 & 58.18 & 4.25 \\
TerraMindv1-L~\cite{TerraMind} & Large    & 82.93        & \textbf{75.57} & \textbf{43.13} & 90.78 & 37.89 & \underline{55.04} & 59.98 & 27.47 & \underline{59.10} & \underline{3.75}\\
SMARTIES-B~\cite{SMARTIES} & Base & 82.80 & & & & 38.50 & & 62.20 & & \\
\midrule
\textbf{RAMEN (ours)} & Base & \textbf{85.02} & \underline{69.72} & \underline{42.29} & \underline{91.03} & \textbf{39.85} & 53.27 & 60.31 & \underline{38.78} & \textbf{60.03} & \textbf{2.63} \\ 
\bottomrule
\end{tabular}
\end{adjustbox}
\label{tab:pangaea-full-results}  
\end{table*}

\subsection{Pretraining data}

To pretrain RAMEN on a wide range of sensors exhibiting varied spatio-spectro-temporal resolutions, we assemble a large-scale EO corpus combining three complementary datasets:
\begin{itemize}
    \item \textbf{FLAIR-HUB}~\cite{FLAIRHUB}:  A multi-sensor land-cover dataset with very high resolution RGB-NIR imagery, 10 bands Sentinel-2 (S2) time series, VV/VH Sentinel-1 (S1) time series and elevation (DSM/DTM) maps covering France. 
    The target GSD is uniformly sampled in the range 3-20\,m at 1\,m interval;
    \item \textbf{WorldStrat}~\cite{WorldStrat}: A global collection covering 10000 $\text{km}^2$ of matched high resolution RGB-NIR imagery and low resolution 12 bands S2 time-series stratified to all type of land-use across the world. 
    To reduce memory constraints, the original SPOT6 (1052x1052px) and S2 (156x156px) tiles are divided in four non-overlapping tiles during pretraining. The target GSD is uniformly sampled in the range 5-20\,m at 1\,m interval;
    \item \textbf{MMEarth64}~\cite{MMEarth}: A large corpus of 1.2 million locations distributed around the world combining 13 bands S2 imagery, 8 bands S1 spanning all available ascending/descending polarizations and elevation (DSM/slope) maps.
    For computational efficiency, we sample $60\%$ of all available locations, stratified by biome, as our pretraining set. The target GSD is uniformly sampled in the range 20-100\,m at 10\,m interval. 
\end{itemize}
To mitigate cross-sensors and cross-datasets distribution shifts, all inputs are standardized with per-channel mean and standard deviation values. 

\subsection{Experimental setup}

\noindent\textbf{Pretraining hyperparameters.} 
We pretrain RAMEN using a ViT-Base backbone~\cite{vit} as our encoder and follow the same architectural parameters as the vanilla MAE whenever possible.
We set the number of convolutional experts to $N_{conv} = 4$ for our main experiment and masking ratio $R$ to $75\%$.
In detail, RAMEN is trained for 100 epochs on a cluster of 16 H100 GPUs using the AdamW optimizer and a base learning rate of $1.5e-4$. 
The learning rate is warmed up for 20 epochs and decayed with a cosine schedule. 
To fit memory requirements, batch size is tailored to the size of input tiles and target GSD range for each dataset: FLAIR-HUB uses a batch size of 64, WorldStrat has a batch size of 32 and MMEarth uses a batch size of 512.

\noindent\textbf{Evaluation protocol.} 
We evaluate RAMEN across 8 downstream datasets from the PANGAEA benchmark~\cite{PANGAEA}. 
It covers a diverse set of unimodal, multimodal and multitemporal EO semantic segmentation tasks including land-cover mapping, flood detection, or wildfire detection.
These datasets span a wide range of modalities (aerial, multispectral, SAR), spatial resolutions (0.2m to 30m), and temporal dynamics (single-date and time-series), providing a rigorous test of model generalization capabilities.
Following TerraMind~\cite{TerraMind}, we exclude xView2 and BioMassters tasks due to non-reproducible results, and omit FiveBillionPixels as the custom dataset and splits used are not available.

For fair comparison, we follow the standardized PANGAEA evaluation protocol on all datasets. 
During experiments, the pretrained RAMEN encoder is frozen, and only a UPerNet~\cite{UPerNet} decoder is finetuned on the downstream tasks. 
To assess the effect of controllable feature map resolution, we conduct extensive experiments varying the target GSD at inference, examining the trade-off between spatial precision, computational cost, and task performance. We report the results obtained by maximizing the mIoU on the validation set of each downstream task.
For large input images, we use the PANGAEA protocol in which random cropping is applied during training to fit the model’s memory constraints.
Multimodal inputs are fused by concatenating features from all modalities along the embedding dimension.

\begin{figure*}[!t]
    \centering
    \includegraphics[width=\textwidth]{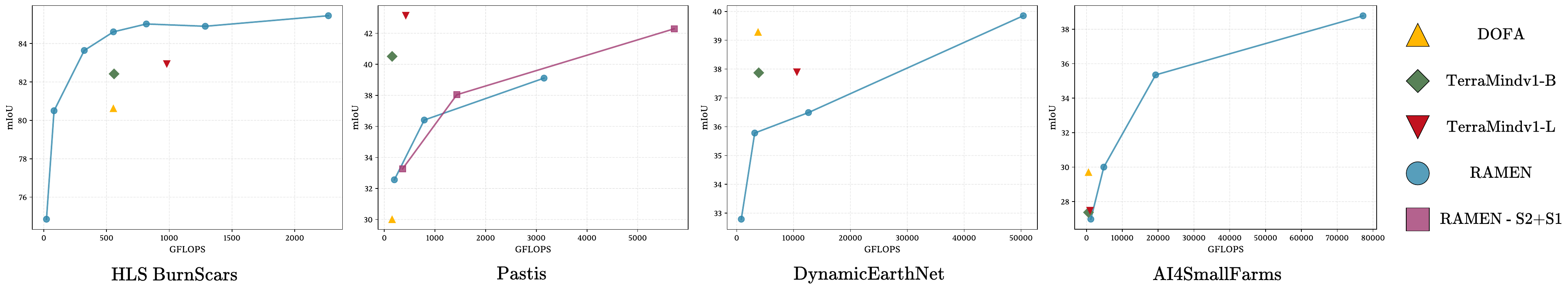}
    \caption{\textbf{Compute/performance trade-off across four downstream tasks.} We plot mIoU versus average GFLOPs per test tile for RAMEN at various target spatial resolutions compared to fixed-resolution foundation models.
    }
    \label{fig:plot_computes}
\end{figure*}

\subsection{Results and analysis}

Tab.~\ref{tab:pangaea-full-results} compares RAMEN against state-of-the-art EO foundation models and baselines across eight diverse semantic segmentation tasks from PANGAEA.
RAMEN sets new state-of-the-art performance, achieving the best average $\text{mIoU}$ of $\textbf{60.03}$ and the best average ranking of $\textbf{2.63}$. Notably, RAMEN surpasses TerraMindv1-L~\cite{TerraMind} ($59.10$ average mIoU) while using a lighter ViT-Base encoder. This illustrates that our resolution-adjustable design and multimodal pretraining strategy enable cross-sensors generalization. 
Beyond overall performance, RAMEN stands out for its consistency across all tasks. It ranks in the top 2 models on 6 out of 8 tasks, highlighting its robustness to diverse EO modalities. We argue that this consistency is enabled by RAMEN ability to adapt its resolution to the specific needs of each task.

\noindent\textbf{Adapting resolution to the task.}
A key advantage of RAMEN is the ability to adjust the feature map resolution during inference through control of the target GSD.
While increasing resolution generally leads to better performance, a crucial finding is that optimal $\text{GSD}_{target}$ is task-dependent and does not always correspond to the finest resolution. 
Tab.~\ref{tab:resolution-specific} illustrates how varying $\text{GSD}_{target}$ impacts performances on two representative tasks:
\begin{itemize}
    \item On HLS BurnScars~\cite{jakubik2023foundation}, a wildfire segmentation task, coarser resolutions yield higher accuracy, as burned areas span large homogeneous regions;
    \item On MADOS~\cite{kikaki2024detecting}, a marine pollutants segmentation task, finer resolutions significantly improve performances, as the model can resolve smaller details.  
\end{itemize}
This adjustable mechanism enables task-aware control over the trade-off between spatial precision and efficiency. In practice, it allows a single model to handle a wide array of real-world applications, from rapid disaster response at coarse scales (e.g. wildfire or flood detection) to fine-grained monitoring (e.g. agriculture delineation, urban mapping, pollutant detection).

\begin{table}[!t]
\centering
\vspace{-1em}
\caption{\textbf{Effects of adjusting $\text{GSD}_{target}$ on performances.} We report the \textbf{validation} mIoU obtained on HLS BurnScars and MADOS datasets for different $\text{GSD}_{target}$ values.}
\begin{adjustbox}{width=1\linewidth}
\begin{tabular}{cccc|cccc}
\toprule
\multicolumn{4}{c}{HLS BurnScars} & \multicolumn{4}{c}{MADOS} \\
Modality & $\text{GSD}_m$ & $\text{GSD}_{target}$ & mIoU & Modality & $\text{GSD}_m$ & $\text{GSD}_{target}$ & mIoU \\
\midrule
\multirow{4}{*}{HLS} & \multirow{4}{*}{30} & 360 & 87.07 & \multirow{4}{*}{S2} & \multirow{4}{*}{10} & 80 & 57.09 \\
& & 300 & \textbf{88.44} & & & 40 & 67.05 \\ 
& & 240 & 88.30 & & & 20 & 76.27 \\ 
& & 180 & 88.09 & & & 10 & \textbf{78.07} \\ 
\bottomrule
\end{tabular}
\end{adjustbox}
\label{tab:resolution-specific}  
\vspace{-1em}
\end{table}

\noindent\textbf{Compute/performance trade-off.}
Fig.~\ref{fig:plot_computes} illustrates the compute/performance trade-off of RAMEN on four representative datasets, showing how resolution adjustability enables direct control over the balance between accuracy and computational cost. As a transformer-based architecture, RAMEN exhibits quadratic complexity with the number of tokens, causing GFLOPs to increase rapidly at finer resolutions. Yet across tasks RAMEN remains competitive with fixed-resolution baselines at comparable computational budgets — achieving $85.02$ mIoU at $817$ GFLOPs on BurnScars (versus TerraMind-L $82.93$ mIoU at $980$ GFLOPs) and reaching $35.78$ mIoU at $3170$ GFLOPs on DynamicEarthNet compared to DOFA ($39.29$ mIoU at $3760$ GFLOPs) and TerraMind-L ($37.89$ mIoU at $10591$ GFLOPs).
Practitioners can select coarser GSDs to reduce costs — on Pastis, $33.26$ mIoU at $359$ GFLOPs achieves $\sim80\%$ of peak performance with $\sim7.4\times$ faster inference.
Notably, accessing finer resolutions enables RAMEN to surpass fixed-resolution limits: on AI4SmallFarms, RAMEN achieves $38.78$ mIoU where all foundation models plateau below $\sim30$ mIoU, demonstrating that resolution adjustability enables qualitatively better performance on detail-critical tasks.

\noindent\textbf{Multimodal fusion.}
\begin{table}[!t]
\centering
\vspace{-1em}
\caption{\textbf{Multimodal fusion performances.} We report the best \textbf{validation} mIoU across three tasks in both uni- and multimodal setup.}
\begin{adjustbox}{width=1\linewidth}
\begin{tabular}{c|ccc|c}
\toprule
Modalities & Sen1Floods11 & Pastis & CTM-SS & Avg mIoU\\
\midrule
S2 & 89.96 & 40.99 & 57.20 & 62.72 \\
S2+S1 & \textbf{91.20} & \textbf{44.25} & \textbf{57.35} & \textbf{64.27}\\
\bottomrule
\end{tabular}
\end{adjustbox}
\label{tab:mm-comp}  
\vspace{-1em}
\end{table}
RAMEN unified framework naturally handles multimodal inputs by projecting them in a unified resolution-aware latent space.
Tab.~\ref{tab:mm-comp} demonstrates the effectiveness of this approach across three tasks requiring multimodal integration.
Incorporating Sentinel-1 SAR data alongside Sentinel-2 imagery consistently improves performances across all evaluated tasks.
By learning shared representations that embed physical properties of each modality, RAMEN enables effective multimodal fusion without requiring modality-specific architectural adaptations. 

\subsection{Ablations}

\begin{table*}[!t]
\centering
\scriptsize
\caption{\textbf{Ablation studies on RAMEN's resolution-aware modules.}}
\label{tab:ablations}

\vspace{-0.2cm}

\begin{subtable}[t]{1\textwidth}
\centering
\caption{\textbf{Impact of central wavelength channel encoding on performance.} 
Progressively refining the encoded central wavelength from coarse approximations to true Sentinel-2 wavelengths improves mIoU on the Sen1Floods11 dataset. \textbf{Bold} values denote the actual Sentinel-2 band central wavelengths.}
\label{tab:spectral-ablation}
\begin{adjustbox}{width=1\textwidth}
\begin{tabular}{l|c|ccc|ccc|ccc|ccc|c}
\toprule
\multirow{2}{*}{Strategy} & \multicolumn{13}{c}{Encoded band central wavelengths (nm)} & mIoU \\
& CA & R & G & B & RE1 & RE2 & RE3 & NIR-B & NIR-N & H$_2$O & SWIR1 & SWIR2 & SWIR3  \\ 
\midrule
RGB & 490 & \textbf{490} & \textbf{560} & \textbf{665} & 665 & 665 & 665 & 665 & 665 & 665 & 665 & 665 & 665 & 79.14 \\
RGB-NIR & \textbf{440} & \textbf{490} & \textbf{560} & \textbf{665} & 865 & 865 & 865 & 865 & \textbf{865} & 865 & 865 & 865 & 865 & 79.82 \\
Key groups & \textbf{440} & \textbf{490} & \textbf{560} & \textbf{665} & 740 & \textbf{740} & 740 & 865 & \textbf{865} & 865 & 1610 & \textbf{1610} & 1610 & 82.24 \\
Full & \textbf{440} & \textbf{490} & \textbf{560} & \textbf{665} & \textbf{705} & \textbf{740} & \textbf{783} & \textbf{842} & \textbf{865} & \textbf{945} & \textbf{1373} & \textbf{1610} & \textbf{2200} & \textbf{83.74} \\
\bottomrule
\end{tabular}
\end{adjustbox}
\end{subtable}

\vspace{0.2cm}

\begin{subtable}[t]{0.48\textwidth}
\centering
\caption{\textbf{Effects of adjustable resampling.} We report the mIoU on the Sen1Floods11 dataset at four coarse to fine $\text{GSD}_{target}$.}
\label{tab:spatial-ablation}
\begin{adjustbox}{width=1\linewidth}
\begin{tabular}{cc|ccc}
\toprule
$\text{GSD}_m$ & $\text{GSD}_{target}$ & Naive & False encoding & Adjustable \\
\midrule
\multirow{4}{*}{10} & 320 & 77.67 & 78.63 & \textbf{78.84}\\
& 160 & 80.84 & 83.23 & \textbf{83.74}\\
& 80 & 85.45 & 86.02 & \textbf{87.13}\\
& 40 & 88.30 & 88.98 & \textbf{88.99}\\
\bottomrule
\end{tabular}
\end{adjustbox}
\end{subtable}
\hfill
\begin{subtable}[t]{0.48\textwidth}
\centering
\caption{\textbf{Temporal integration performances.} We compare RAMEN with multi-temporal models on three tasks. Late fusion indicates that each timestep is processed independently before aggregation via a LTAE~\cite{LTAE}.}
\label{tab:temporal-comp}
\begin{adjustbox}{width=1\linewidth}
\begin{tabular}{l|ccc|c}
\toprule
Model & Pastis & DEN & CTM-SS & Avg mIoU\\
\midrule
Prithvi & \textbf{33.93} & 27.86 & 43.07 & 34.95 \\
SatlasNet & 17.51 & \textbf{36.31} & 46.97 & 33.59 \\
RAMEN & 28.07 & 33.16 & \textbf{53.01} & \textbf{38.08}\\
\midrule
RAMEN-Late fusion & 42.29 & 39.85 & 53.27 & 45.14 \\
\bottomrule
\end{tabular}
\end{adjustbox}
\end{subtable}

\end{table*}

\noindent\textbf{Wavelength encoding.}
To validate that incorporating physical information improves over treating spectral bands as generic channels, we progressively refine the encoded central wavelengths from coarse approximations to true Sentinel-2 central wavelengths.
We test four encoding strategies with increasing spectral precision: from encoding only RGB channels while collapsing others to red wavelength, to progressively distinguishing RGB-NIR, key spectral groups (Coastal Aerosol, RGB, Red-Edge, NIR, SWIR), and finally using true central wavelengths for all 13 bands.

Tab.~\ref{tab:spectral-ablation} shows results on the Sen1Floods11 dataset using the 13 Sentinel-2 bands. 
We observe consistent improvements with increasing spectral precision, with a total gain of $+4.6\,\text{mIoU}$ from the RGB-only true encoding to full-precision. This improvement demonstrates that incorporating the physical meaning of spectral bands rather than treating them as abstract feature channels enables the model to learn more semantically meaningful representations.

\noindent\textbf{Spatial resampling module.}
To evaluate the benefit of our adjustable spatial resampling strategy, we compare three variants of RAMEN differing only in how we handle resampling of the input data to the target GSD at inference.
(1) \textbf{Naive resampling}: inputs are directly bilinearly interpolated without convolutional adaptation;
(2) \textbf{False encoding}: inputs are resampled but a false interpolation ratio (set to 1) is provided to the spatial encoding. This prevents our mixture of convolutional experts from adapting to the true interpolation scale;
(3) \textbf{Adjustable resampling}: inputs are resampled with the true interpolation ratio encoded (i.e. using our proposed spatial resampler).

As shown in Tab.~\ref{tab:spatial-ablation}, our adjustable resampling strategy consistently improves mIoU across all target GSDs, demonstrating that explicitly encoding the interpolation ratio allows the mixture of convolutional experts to adapt effectively to varying resolutions. The benefit of adjustable resampling becomes increasingly pronounced as the target GSD moves further from the original GSD.

\noindent\textbf{Temporal attention.}
To ensure fair comparison with non-temporal models that handle time-series via learned late fusion during fine-tuning, we evaluate RAMEN under both our native temporal attention and late fusion configuration.
Tab.~\ref{tab:temporal-comp} shows results across three tasks with diverse temporal dynamics — from first 6 days of every month for DEN to evenly distributed over years samples for Pastis.
Among multi-temporal models, Prithvi achieves $34.95$ average mIoU and SatlasNet $33.59$, while RAMEN reaches $38.14$.
When using late fusion, performance gains are highly task-dependent: CTM-SS improves minimally (+$0.26$ mIoU), while Pastis and DEN gain substantially (+$14.02$ and +$6.69$ respectively).
These results show that a late fusion provides additional gains depending on temporal dynamics inherent to each dataset, at the cost of fine-tuning an additional LTAE module. However, RAMEN temporal handling, without the need to fine-tune an additional module, outperforms other specialized temporal models.

\section{Conclusion}
We presented RAMEN, a resolution-adjustable encoder for EO that enables practitioners to dynamically control feature map resolution at inference time to match task-specific requirements.
To the best of our knowledge, RAMEN is the first EO foundation model which can simultaneously learn modality-agnostic, multi-temporal and resolution-adjustable representations.
Once pretrained, our model sets new state-of-the-art performances across multiple tasks and modalities and outperforms larger models.
This framework provides a scalable foundation for general-purpose EO models capable of adapting to diverse sensor configurations and application needs.

\section*{Acknowledgement}
This project was supported by ANR project ANR-23-IAS1-0002 GEO ReSeT and provided with computing AI and storage resources by GENCI at IDRIS thanks to the grant 2025-AD011016746 on the supercomputer Jean Zay's H100 partition.

{
    \small
    \bibliographystyle{ieeenat_fullname}
    \bibliography{main}
}

\clearpage
\setcounter{page}{1}
\renewcommand\thefigure{\Alph{figure}}
\renewcommand\thesection{\Alph{section}}
\renewcommand\thetable{\Alph{table}}
\renewcommand\theequation{\Alph{equation}}
\setcounter{equation}{0}
\setcounter{section}{0}
\setcounter{figure}{0}
\setcounter{table}{0}
\maketitlesupplementary

In the supplementary, we provide implementation details about RAMEN pretraining and evaluation (Sec.~\ref{sup:impl-details}), 
additional analysis of RAMEN architecture and efficiency (Sec.~\ref{sup:analysis}),
extended results on all downstream tasks in PANGAEA benchmark (Sec.~\ref{sup:full-results}) and qualitative examples of RAMEN representations at various $\text{GSD}_{target}$ (Sec.~\ref{sup:qualitative}).

\section{Implementation details}
\label{sup:impl-details}

\subsection{RAMEN architecture}

RAMEN follows the Vision Transformer architecture \cite{vit} and has $98.5M$ learnable parameters, all shared across sensors except for the channel-conditioned projectors separated across three modality types (Optical, radar and DEM):
\begin{itemize}
    \item \textbf{Channel-conditioned projectors ($3\times2.4M$ parameters).}
    Each channel-conditioned projector consists of a MLP processing channel-conditioned encodings to produce a channel-wise projection matrix;
    
    \item \textbf{Spatial resampler ($2.5M$ parameters).}
    Composed of $4$ $1\times1$ convolution experts. A MLP and softmax layer process log scaled interpolation ratio encoding to produce normalized weights $\{w_n\}_{n=1}^{4}$;
    
    \item \textbf{Temporal encoder ($3.7M$ parameters).} 
    A lightweight temporal attention encoder (LTAE) \cite{LTAE} processes time-series inputs. Sinusoidal encoding based on day of acquisition is added to each timestep before temporal aggregation;
   
    \item \textbf{Encoder blocks ($85.1M$ parameters).} 
    We follow the standard ViT-Base architecture composed of $12$ self-attention blocks with $12$ heads. The embedding dimension $D$ is set to $768$.
\end{itemize}

\noindent\textbf{Decoder blocks and reconstruction ($33.0M$ parameters).}
We follow the MAE \cite{mae} framework for the decoder architecture. 
In details, the decoder has an embedding dimension of $512$ and is composed of $8$ self-attention blocks with $16$ heads. 
Resolution-adjustable reconstruction modules are similar to their corresponding projection ones, except for the temporal reconstruction composed of one self-attention block processing feature map expanded and enriched with day-of-acquisition encoding independently. 

\noindent\textbf{GSD-based positional encoding.}
To ensure coherent processing across arbitrary target resolutions, RAMEN incorporates GSD-based positional encodings following Scale-MAE~\cite{Scale-MAE}. 
These encodings embed the target GSD directly into the sinusoidal positional functions as:
\begin{align}
    \text{GSDPE}(\text{pos}_x, 2k) &= \sin\left(\frac{\text{GSD}_{target}}{G}\frac{\text{pos}_x}{10000^{2k/D}}\right)\,; \\
    \text{GSDPE}(\text{pos}_y, 2k+1) &= \cos\left(\frac{\text{GSD}_{target}}{G}\frac{\text{pos}_y}{10000^{2k/D}}\right)\,,
\end{align}
where $G$ is a reference length set to one.

This formulation is essential for our resolution-adjustable behavior: 
while RAMEN can ingest any number of tokens depending on $\text{GSD}_{target}$, the positional encoding ties the representation to a consistent physical scale across coarse and fine resolutions. 
This enables the encoder to maintain spatial coherence and interpret feature maps in a resolution-aware manner.

\subsection{RAMEN pretraining corpus}

RAMEN pretraining data covers a wide range of heterogeneous EO modalities, incorporating diverse modal, spatial and temporal resolutions.
See Tab.~\ref{tab:data_corpus} for more details on RAMEN pretraining corpus characteristics.

\begin{table*}[!t]
    \centering
    \caption{\textbf{RAMEN pretraining corpus.} 
    We combine three large-scale multimodal EO datasets covering diverse spectral, spatial and temporal resolutions. The image size is expressed in pixels for a square image.
    During pretraining, the target GSD  is randomly sampled from a dataset-specific range.
    }
    \begin{adjustbox}{width=1\textwidth}
    \begin{tabular}{lccccccccc}
        \toprule
        Dataset & Spatial extent & Num tiles & Modality & Image size & Bands & Time-series & GSD (m) & Target GSD range (interval) & Batch size (pretraining)\\
        \midrule
        \multirow{4}{*}{FLAIR-HUB~\cite{FLAIRHUB}} & \multirow{4}{*}{2 528 $\text{km}^2$} & \multirow{4}{*}{241 100} & Aerial VHR & 512 & 4 & & 0.2 & \multirow{4}{*}{3 - 20m (1m)} & \multirow{4}{*}{64}\\
        & & & S2 & 10 & 10 & \cmark & 10 & \\
        & & & S1 & 10 & 2 & \cmark & 10 &\\
        & & & DEM & 512 & 2 & & 0.2 & \\
        \midrule
        \multirow{2}{*}{WorldStrat~\cite{WorldStrat}} & \multirow{2}{*}{9 820 $\text{km}^2$} & \multirow{2}{*}{62 848} & SPOT6 & 263 & 4 & & 1.5 & \multirow{2}{*}{5 - 20m (1m)} & \multirow{2}{*}{32}\\
        & & & S2 & 39 & 12 & \cmark & 10 & \\
        \midrule
        \multirow{3}{*}{MMEarth64~\cite{MMEarth}} & \multirow{3}{*}{460 800 $\text{km}^2$} & \multirow{3}{*}{720 000} & S2 & 64 & 13 & & 10 & \multirow{3}{*}{20 - 100m (10m)} & \multirow{3}{*}{512}\\
        & & & S1 & 64 & 8 & & 10 &\\
        & & & DEM & 64 & 2 & & 10 &\\
        \bottomrule
    \end{tabular}
    \end{adjustbox}
    \label{tab:data_corpus}
\end{table*}

\subsection{PANGAEA evaluation protocol}

For all downstream tasks, we follow the standardized PANGAEA~\cite{PANGAEA} evaluation protocol. 
In details, the pretrained encoder is frozen while a UPerNet~\cite{UPerNet} decoder is finetuned for 80 epochs.
On all tasks, AdamW optimizer is used with a base learning rate of $1e-4$, weight decay of 0.05 and batch size of 8. 
The learning rate is decayed 10$\times$ after $60\%$ and $90\%$ of the total steps.
On multi-temporal tasks, non-temporal models and RAMEN - Late fusion process each timestep independently before aggregation with a lightweight temporal attention encoder (LTAE) \cite{LTAE}.

To process large input tiles, random cropping is applied during training to match the encoder expected input size. 
For evaluation, a sliding window inference strategy is employed, dividing the image in evenly distributed smaller crops.
While RAMEN can natively handle any input size, we restricted input size on large tiles and high resolutions experiments to match memory and time-processing constraints. Extensive results for diverse input size and resolutions can be found in Sec.~\ref{sup:full-results}.
We refer the readers to PANGAEA~\cite{PANGAEA} for detailed informations on the evaluation protocol.

\subsection{GFLOPs and inference time calculation}

We use the \emph{fvcore} \cite{fvcore} library to compute GFLOPs estimation of processing one input tile for each dataset. 
We include for these estimations the processing time of the encoder and UPerNet decoder.
Because of PANGAEA sliding window inference strategy, we multiply GFLOPs obtained for one crop by the number of cropped inputs necessary to produce the final segmentation map.
Average inference time per tile is computed over 10 steps.
All results of GFLOPs and inference time per task for model, input size and $\text{GSD}_{target}$ are reported in Sec.~\ref{sup:full-results}.

\section{Additional analysis}
\label{sup:analysis}

\subsection{Generalization to unseen sensors}

\begin{table}[!t]
\centering
\caption{\textbf{Unseen sensor performances.} We report the mIoU and rank (across all PANGAEA models) on  three tasks using sensors unseen during pretraining. 
RAMEN achieves the highest average mIoU and rank, highlighting the robustness of its modality-agnostic design and its ability to transfer to new sensor configurations.}
\label{tab:unseen-comp}
\begin{adjustbox}{width=1\linewidth}
\begin{tabular}{l|ccc|c}
\toprule
\multirow{2}{*}{Model} & \multicolumn{3}{c}{mIoU (Rank)} & \multirow{2}{*}{Avg. mIoU (Rank)} \\
 & BurnSr & DEN & SN7 & \\
\midrule
CROMA & 82.42 (5) & 38.29 (4) & 59.28 (9) & 60.00 (6.00)\\
DOFA & 80.63 (12) & 39.29 (3) & \textbf{61.84} (4) & 60.59 (6.33)\\
Terramindv1-B & 82.42 (5) & 37.87 (6) & 60.61 (6) & 60.30 (5.66) \\
Terramindv1-L & 82.93 (4) & 37.89 (5) & 59.98 (8) & 60.27 (5.66) \\
\midrule
RAMEN & \textbf{85.02} (1) & \textbf{39.85} (1) & 60.31 (7) & \textbf{61.73} (\textbf{3.00})\\
\bottomrule
\end{tabular}
\end{adjustbox}
\end{table}

Tab.~\ref{tab:unseen-comp} reports performance on three tasks featuring sensors unseen during pretraining. RAMEN achieves the highest average mIoU and rank across the benchmark, notably outperforming all other foundation models on BurnScars and DynamicEarthNet, two tasks exhibiting widely different sensor configurations (HLS - $6$ multispectral channels/$30$m GSD and PlanetFusion - RGB-NIR/$3$m GSD respectively).

These results validate the sensor-agnostic design of RAMEN. 
The channel-conditioned projector adapts to per-channel characteristics, while our resolution-adjustable framework provides a consistent interface across modalities and spatial scales.
As a result, RAMEN transfers effectively to new sensors without requiring sensor-specific pretraining or architectural adaptation, demonstrating strong robustness on heterogeneous EO data.

\subsection{Spatial resampling analysis}

\begin{figure}[!t]
    \centering
    \includegraphics[width=\linewidth]{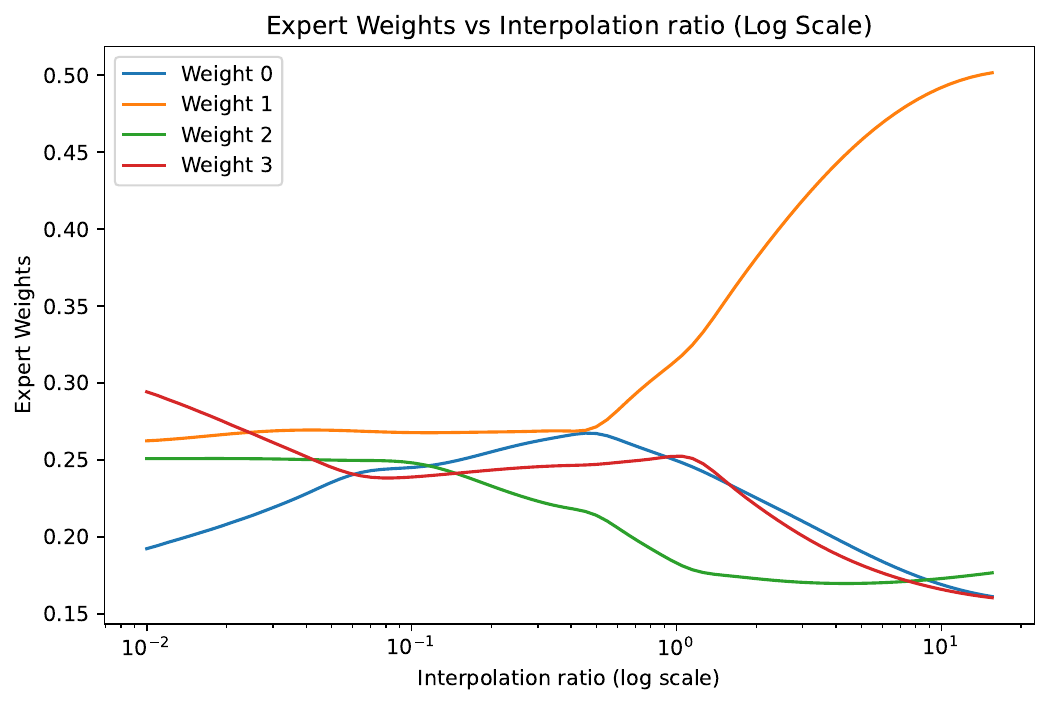}
    \caption{\textbf{Specialization of the convolutional experts across interpolation ratios}. We plot the normalized weights of the four convolutional experts as a function of the interpolation ratio $\text{GSD}_m/\text{GSD}_{target}$.
    }
    \label{fig:weightsplot}
\end{figure}

In standard Vision Transformer architectures, input images are divided into fixed-size patches using strided convolutions (e.g., $16\times16$ convolution with stride 16). 
However, this method does not let us explicitly control the resolution of the resulting feature map, nor does it allow the model to understand the ground sampling distance associated with each pixel.

Our spatial resampling module takes a different approach through a two-stage process. 
First, bilinear interpolation geometrically aligns features to the target GSD, establishing consistent spatial structure across heterogeneous inputs. 
This interpolation handles all spatial resampling. 
Second, a mixture of $1\times1$ convolutional experts refines the interpolated features through scale-specific transformations.
The use of $1\times1$ kernels for our convolution experts is central to our design philosophy: they operate independently on each spatial location, adjusting channel statistics without introducing any spatial dependencies. 
After interpolation has established the spatial structure, the experts focus purely on feature-space refinement and adapting channel responses based on the interpolation ratio.

\noindent\textbf{Expert specialization across interpolation scales.}
Fig.~\ref{fig:weightsplot} reveals how the four convolutional experts specialize across different interpolation ratios $\text{GSD}_m / \text{GSD}_{target}$. 
While the weights vary smoothly with the scale factor, we note some clear specialization patterns:

\begin{itemize}
    \item \textbf{Expert 1 handle upsampling regimes.}
    When $\text{GSD}_m/\text{GSD}_{target}>1$ (i.e., when RAMEN increases spatial resolution), Expert 1 consistently receives the highest weight. 
    This indicates a learned specialization for refining interpolated coarse inputs;
    
    \item \textbf{Balanced mixture near identity and weak downsampling.}
    Around $\text{GSD}_m/\text{GSD}_{target} \sim 1$ down to $0.1$, the weights of the experts remain relatively even. 
    In this regime, the model appears to rely on a mixture of all transformations rather than favoring any single convolution;
    
    \item \textbf{Expert 3 dominates under strong downsampling.}
    For heavy downsampling ratios (up to $10e-2$), the weights gradually shift toward another layer (Expert 3), while Expert 0 become less prominent. 
    This suggests that certain transformations are more suitable when the input has been heavily compressed and requires coarser adjustments.
\end{itemize}

\noindent\textbf{Additional ablations on adjustable resampling.}
Tab.~\ref{tab:spatial-ablation-rebuttal} reports adjustable resampling effect compared with n aive and false encoding strategy on two additional tasks, with an \textbf{average gain of 0.48 mIoU} across the three evaluated tasks, demonstrating the added benefit of our scale-conditioned refinement of interpolated features.

\begin{table}[t]
\centering
\caption{\textbf{Effects of adjustable resampling.} We report the validation mIoU for coarse to fine
$\text{GSD}_{target}$.}
\label{tab:spatial-ablation-rebuttal}
\begin{adjustbox}{width=1\linewidth}
\begin{tabular}{ccc|ccc}
\toprule
Task & $\text{GSD}_m$ & $\text{GSD}_{target}$ & Naive & False encoding & Adjustable \\
\midrule
\multirow{4}{*}{HLS BurnScars} & \multirow{4}{*}{30} & 1920 & 75.88 & 76.42 & \textbf{77.14} \\
& & 960 & 79.70 & 81.87 & \textbf{82.45} \\
& & 480 & 83.98 & 86.31 & \textbf{87.07} \\
& & 300 & 86.51 & 87.64 & \textbf{88.44} \\
\midrule
\multirow{3}{*}{Pastis (S2)} & \multirow{3}{*}{10} & 80 & 29.42 & 32.92 & \textbf{32.97} \\
& & 40 & 32.22 & 37.85 & \textbf{38.02} \\
& & 20 & 37.35 & 40.66 & \textbf{40.99} \\
\bottomrule
\end{tabular}
\end{adjustbox}
\end{table}

\subsection{Pretraining efficiency}

RAMEN was pretrained for 100 epochs in approximately 50 hours on a cluster of 16 H100 GPUs cluster, corresponding to $\approx800$ GPU-hours.
This represents a considerably lower computational budget than that reported by other EO foundation models.
Notably, TerraMind~\cite{TerraMind} reports a pretraining cost of about 6 days on 32 A100s GPUs for its Base version ($\approx$ 4608 GPU-hours) and over 10 days for its Large configuration ($\approx$ 7680 GPU-hours).

\section{Detailed results}
\label{sup:full-results}

\begin{figure*}[!t]
    \centering
    \includegraphics[width=\textwidth]{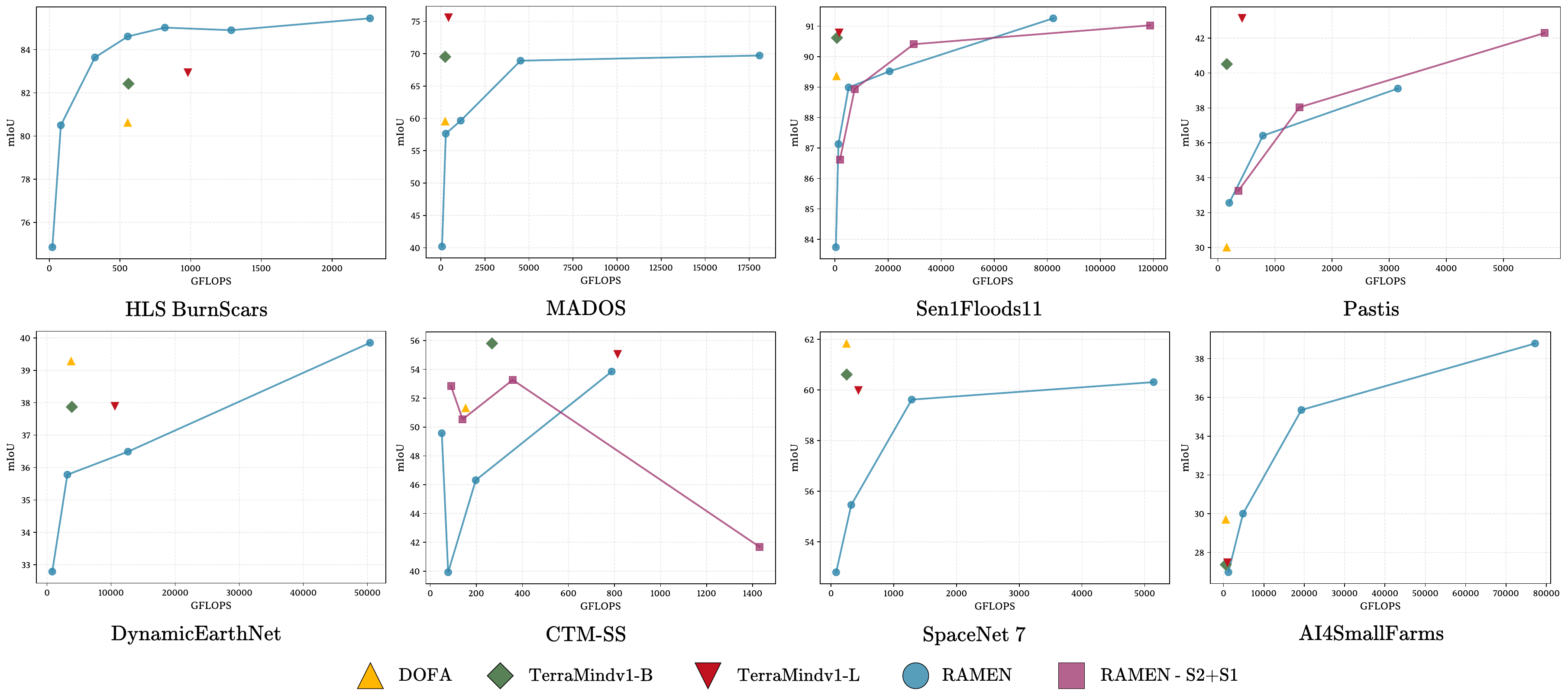}
    \caption{\textbf{Compute/performance trade-off across eight downstream tasks.} We plot mIoU versus average GFLOPs per test tile for RAMEN at various target spatial resolutions compared to fixed-resolution foundation models.
    }
    \label{fig:plot_computes_full}
\end{figure*}

We present in this section extended results across various $\text{GSD}_{target}$ on all downstream tasks. Fig.~\ref{fig:plot_computes_full} illustrates the compute/performance trade-off on the eight downstream tasks considered. 

\subsection{HLS BurnScars (BurnSr)}

HLS BurnScars~\cite{jakubik2023foundation} is a post-fire burn scar segmentation task using Harmonized Landsat-Sentinel
(HLS) imagery. 
It contains $804$ tiles of size $512\times512$ at $\text{GSD}=30m$, with six multispectral bands (RGB-NIR-SWIR1-SWIR2). 

Tab.~\ref{tab:burnscars} presents the obtained results, GFLOPs estimation and inference time per tile across various $\text{GSD}_{target}$.

\subsection{MADOS}

Marine Debris and Oil Spill (MADOS)~\cite{kikaki2024detecting} is a marine pollutants segmentation task using Sentinel-2 data. It contains $2803$ tiles of size $240\times240$ at $\text{GSD}=10m$,with eleven multispectral bands.

Tab.~\ref{tab:MADOS} presents the obtained results, GFLOPs estimation and inference time per tile across various $\text{GSD}_{target}$.

\subsection{Pastis}

Pastis~\cite{garnot2021panoptic} is a semantic segmentation task of agricultural parcels using multi-temporal Sentinel-2 and Sentinel-1 data. It contains $2433$ tiles of size $128\times128$ at $\text{GSD}=10m$, with ten multispectral bands and two polarizations (VV/VH) respectively.
Following PANGAEA benchmark setup, $6$ evenly distributed over time captures are selected.

Tab.~\ref{tab:Pastis} and Tab.~\ref{tab:Pastis-LTAE} detail the obtained results, GFLOPs estimation and inference time per tile across various $\text{GSD}_{target}$ using RAMEN temporal encoder and Late LTAE fusion respectively.

\subsection{Sen1Floods11 (Sen1Fl11)}
Sen1Floods11~\cite{rambour2020flood} is a global flood mapping segmentation task using Sentinel-2 and Sentinel-1 data. It contains $4831$ tiles of size $512\times512$ at $\text{GSD}=10m$, with $13$ multispectral bands and two polarizations (VV/VH) respectively.

Tab.~\ref{tab:Sen1Floods11} presents the obtained results, GFLOPs estimation and inference time per tile across various $\text{GSD}_{target}$.

\subsection{DynamicEarthNet (DYN)}
DynamicEarthNet~\cite{toker2022dynamicearthnet} is a semantic segmentation task using multi-temporal at daily observations PlanetFusion data. It spans $75$ areas of interest of size $1024\times1024$ at $\text{GSD}=3m$, with RGB-NIR bands.
Following PANGAEA benchmark setup, the $6$ first captures of every month are selected.

Tab.~\ref{tab:dyn_no_ltae} and Tab.~\ref{tab:dyn_ltae} detail the obtained results, GFLOPs estimation and inference time per tile across various $\text{GSD}_{target}$ using RAMEN temporal encoder and Late LTAE fusion respectively.

\subsection{CropTypeMapping - South Sudan (CTM-SS)}

CropTypeMapping - South Sudan~\cite{m2019semantic} is a semantic segmentation task of agricultural parcels in underrepresented regions using multi-temporal Sentinel-2 and Sentinel-1 data. 
It contains $837$ tiles of size $64\times64$ at $\text{GSD}=10m$, with ten multispectral bands and two polarizations (VV/VH) respectively.

Tab.~\ref{tab:CTM_no_ltae} and Tab.~\ref{tab:CTM_ltae} detail the obtained results, GFLOPs estimation and inference time per tile across various $\text{GSD}_{target}$ using RAMEN temporal encoder and Late LTAE fusion respectively.

\subsection{SpaceNet 7 (SN7)}

SpaceNet 7~\cite{van2018spacenet} is a urban semantic segmentation task using PlanetScope data. 
It contains $15973$ tiles of size $256\times256$ at $\text{GSD}=3m$, with RGB-NIR bands.

Tab.~\ref{tab:SpaceNet7} presents the obtained results, GFLOPs estimation and inference time per tile across various $\text{GSD}_{target}$.

\subsection{AI4SmallFarms (AI4Farms)}

AI4SmallFarms~\cite{ai4small} is an agricultural delineation semantic segmentation task using Sentinel-2 data. 
It contains $62$ tiles of size $496\times496$ at $\text{GSD}=10m$, with RGB-NIR bands.
As no validation set is given in PANGAEA benchmark, we computed test scores on the last checkpoint (epoch 80) after training.

Tab.~\ref{tab:ai4smallfarms} presents the obtained results, GFLOPs estimation and inference time per tile across various $\text{GSD}_{target}$.

\begin{table*}[!t] 
\centering 
\small 
\caption{\textbf{HLS BurnScars (BurnSr) performances.} We report validation and test mIoU, along with GFLOPs and inference time per tile for RAMEN evaluated at different $\text{GSD}_{target}$.} 
\begin{adjustbox}{width=1\textwidth} 
\begin{tabular}{lcccc|cccc} 
\toprule 
Model & Img size & $\text{GSD}_m$ & Input size & $\text{GSD}_{target}$ & val mIoU & test mIoU & GFLOPs & Inference time/tile (ms)\\
\midrule 
\multirow{7}{*}{RAMEN} & \multirow{7}{*}{512} & \multirow{7}{*}{30} & \multirow{7}{*}{512} & 1920 & 77.14 & 74.85 & 21.41 & 6.87\\ 
& & & & 960 & 82.45 & 80.50 & 81.67 & 7.96 \\ 
& & & & 480 & 87.07 & 83.64 & 322.68 & 14.75 \\ 
& & & & 360 & 87.54 & 84.61 & 554.91 & 21.13 \\ 
& & & & 300 & \textbf{88.44} & 85.02 & 817.25 & 30.31 \\ 
& & & & 240 & 88.30 & 84.90 & 1286.74 & 51.84 \\ 
& & & & 180 & 88.09 & 85.45 & 2268.13 & 112.67 \\ 
\midrule 
DOFA & & & & & & 80.63 & 554.51 & 29.38\\ 
TerraMind-B & & & & & & 82.42 & 560.12 & 27.37\\ 
TerraMind-L & & & & & & 82.93 & 980.06 & 53.41 \\ 
\bottomrule 
\end{tabular} 
\end{adjustbox} 
\label{tab:burnscars} 
\end{table*}

\begin{table*}[!t]
\centering
\small
\caption{\textbf{MADOS performances.} We report validation and test mIoU, along with GFLOPs and inference time per tile for RAMEN evaluated at different $\text{GSD}_{target}$.} 
\begin{adjustbox}{width=1\textwidth}
\begin{tabular}{lcccc|cccc}
\toprule
Model & Img size & $\text{GSD}_m$ & Input size & $\text{GSD}_{target}$ & val mIoU & test mIoU & GFLOPs & Inference time/tile (ms)\\
\midrule
\multirow{6}{*}{RAMEN} & \multirow{6}{*}{240} & \multirow{6}{*}{10} & \multirow{2}{*}{240} & 160 & 46.59 & 40.18 & 71.16 & 7.13\\
& & & & 80 & 57.09 & 57.64 & 283.15 & 13.38 \\
\cmidrule(lr){4-9}
& & & 120 & 40 & 67.05 & 59.65 & 1131.15 & 37.02 \\
\cmidrule(lr){4-9}
& & & \multirow{2}{*}{60} & 20 & 76.27 & 68.92 & 4523.13 & 139.95 \\
& & & & 10 & \textbf{78.07} & 69.72 & 18084.79 & 600.00 \\
\midrule
DOFA & &  & & & & 59.58 & 247.83 & 13.67 \\
TerraMind-B & & & & & & 69.52 & 249.04 & 12.83 \\
TerraMind-L & & & & & & 75.57 & 435.68 & 24.44 \\
\bottomrule
\end{tabular}
\end{adjustbox}
\label{tab:MADOS}  
\end{table*}

\begin{table*}[!t]
\centering
\small
\caption{\textbf{Pastis with RAMEN temporal encoder performances.} We report validation and test mIoU, along with GFLOPs and inference time per tile for RAMEN evaluated at different $\text{GSD}_{target}$.} 
\begin{adjustbox}{width=1\textwidth}
\begin{tabular}{lccccc|cccc}
\toprule
Model & Modality & Img size & $\text{GSD}_m$ & Input size & $\text{GSD}_{target}$ & val mIoU & test mIoU & GFLOPs & Inference time/tile (ms)\\
\midrule
\multirow{6}{*}{RAMEN} & \multirow{3}{*}{S2} & \multirow{3}{*}{128} & \multirow{3}{*}{10} & \multirow{2}{*}{128} & 80 & 24.07 & 23.58 & 86.26 & 10.30 \\
& & & & & 40 & 26.51 & 25.93 & 342.38 & 17.35 \\
\cmidrule(lr){5-10}
& & & & 64 & 20 & 23.46 & 23.14 & 1367.25 & 52.78 \\
\cmidrule(lr){2-10}
& \multirow{2}{*}{S2+S1} & \multirow{2}{*}{128} & \multirow{3}{*}{10} & 128 & 40 & 28.88 & 28.27 & 504.89 & 29.38 \\
\cmidrule(lr){5-10}
& & & & 64 & 20 & \textbf{29.62} & 28.07 & 2016.87 & 96.98 \\
\midrule 
DOFA & & &  & & & & 30.02 & 153.52 & 27.16\\
TerraMind-B & & & & & & & 40.51 & 154.94 & 21.88\\
TerraMind-L & & & & & & & 43.13 & 423.70 & 41.66\\
\bottomrule
\end{tabular}
\end{adjustbox}
\label{tab:Pastis}  
\end{table*}

\begin{table*}[!t]
\centering
\small
\caption{\textbf{Pastis with late LTAE fusion performances.} We report validation and test mIoU, along with GFLOPs and inference time per tile for RAMEN evaluated at different $\text{GSD}_{target}$.} 
\begin{adjustbox}{width=1\textwidth}
\begin{tabular}{lccccc|cccc}
\toprule
Model & Modality & Img size & $\text{GSD}_m$ & Input size & $\text{GSD}_{target}$ & val mIoU & test mIoU & GFLOPs & Inference time/tile (ms)\\
\midrule
\multirow{6}{*}{RAMEN} & \multirow{3}{*}{S2} & \multirow{3}{*}{128} & \multirow{3}{*}{10} & \multirow{2}{*}{128} & 80 & 32.97 & 32.56 & 198.24 & 32.54 \\
& & & & & 40 & 38.02 & 36.41 & 788.66 & 63.30 \\
\cmidrule(lr){5-10}
& & & & 64 & 20 & 40.99 & 39.11 & 3152.37 & 177.77 \\
\cmidrule(lr){2-10}
& \multirow{3}{*}{S2+S1} & \multirow{3}{*}{128} & \multirow{3}{*}{10} & \multirow{2}{*}{128} & 80 & 35.51 & 33.26 & 358.91 & 50.17 \\
& & & & & 40 & 40.03 & 38.04 & 1430.76 & 112.62 \\
\cmidrule(lr){5-10}
& & & & 64 & 20 & \textbf{44.25} & 42.29 & 5720.32 & 371.49 \\
\midrule 
DOFA & & & & & & & 30.02 & 153.52 & 27.16\\
TerraMind-B & & & & & & & 40.51 & 154.94 & 21.88\\
TerraMind-L & & & & & & & 43.13 & 423.70 & 41.66\\
\bottomrule
\end{tabular}
\end{adjustbox}
\label{tab:Pastis-LTAE}  
\end{table*}

\begin{table*}[!t]
\centering
\small
\caption{\textbf{Sen1Floods11 performances.} We report validation and test mIoU, along with GFLOPs and inference time per tile for RAMEN evaluated at different $\text{GSD}_{target}$.} 
\begin{adjustbox}{width=1\textwidth}
\begin{tabular}{lccccc|cccc}
\toprule
Model & Modality & Img size & $\text{GSD}_m$ & Input size & $\text{GSD}_{target}$ & val mIoU & test mIoU & GFLOPs & Inference time/tile (ms)\\
\midrule
\multirow{9}{*}{RAMEN} & \multirow{5}{*}{S2} & \multirow{5}{*}{512} & \multirow{5}{*}{10} & 512 & 160 & 82.32 & 83.74 & 324.11 & 14.68\\
\cmidrule(lr){5-10}
& & & & 256 & 80 & 86.09 & 87.13 & 1288.57 & 42.12\\
\cmidrule(lr){5-10}
& & & & 128 & 40 & 88.70 & 88.99 & 5146.42 & 153.71\\
\cmidrule(lr){5-10}
& & & & 64 & 20 & 89.82 & 89.52 & 20577.85 & 609.96\\
& & & & 64 & 10 & 89.96 & 91.26 & 82277.51 & 2850.12\\
\cmidrule(lr){2-10}
& \multirow{4}{*}{S2+S1} & \multirow{4}{*}{512} & \multirow{4}{*}{10} & 256 & 80 & 86.43 & 86.61 & 1858.47 & 76.08 \\
\cmidrule(lr){5-10}
& & & & 128 & 40 & 88.94 & 88.94 & 7424.81 & 280.00\\
\cmidrule(lr){5-10}
& & & & 64 & 20 & 90.45 & 90.41 & 29690.20 & 1119.95\\
& & & & 64 & 10 & \textbf{91.20} & 91.03 & 118721.00 & 7539.63\\
\midrule
DOFA & & & & & & & 89.37 & 559.72 & 29.92\\
TerraMind-B & & & & & & & 90.62 & 729.93 & 37.11 \\
TerraMind-L & & & & & & & 90.78 & 1565.03 & 85.63 \\
\bottomrule
\end{tabular}
\end{adjustbox}
\label{tab:Sen1Floods11}  
\end{table*}

\begin{table*}[!t]
\centering
\small
\caption{\textbf{DynamicEarthNet with RAMEN temporal encoder performances.} We report validation and test mIoU, along with GFLOPs and inference time per tile for RAMEN evaluated at different $\text{GSD}_{target}$.} 
\begin{adjustbox}{width=1\textwidth}
\begin{tabular}{lcccc|cccc}
\toprule
Model & Img size & $\text{GSD}_m$ & Input size & $\text{GSD}_{target}$ 
& val mIoU & test mIoU & GFLOPs & Inference time/tile (ms)\\
\midrule
\multirow{4}{*}{RAMEN} & \multirow{4}{*}{1024} & \multirow{4}{*}{3} & \multirow{2}{*}{512}
& 96 & 28.21 & 32.26 & 361.18 & 35.46 \\
& & & & 48 & 29.72 & 33.52 & 1385.32 & 72.75\\
\cmidrule(lr){4-9}
& & & \multirow{2}{*}{256} & 24 & 31.08 & 31.23 & 5483.24 & 216.42 \\
& & & & 12 & \textbf{31.71} & 33.16 & 21869.41 & 870.45\\
\midrule
DOFA & & & & &  & 39.29 & 3760.22 & 230.98\\
TerraMind-B & & & & &  & 37.87 & 3873.00 & 205.79\\
TerraMind-L & & & & &  & 37.89 & 10591.81 & 625.38\\
\bottomrule
\end{tabular}
\end{adjustbox}
\label{tab:dyn_no_ltae}
\end{table*}

\begin{table*}[!t]
\centering
\small
\caption{\textbf{DynamicEarthNet with late LTAE fusion performances.} We report validation and test mIoU, along with GFLOPs and inference time per tile for RAMEN evaluated at different $\text{GSD}_{target}$.} 
\begin{adjustbox}{width=1\textwidth}
\begin{tabular}{lcccc|cccc}
\toprule
Model & Img size & $\text{GSD}_m$ & Input size & $\text{GSD}_{target}$ 
& val mIoU & test mIoU & GFLOPs & Inference time/tile (ms)\\
\midrule
\multirow{4}{*}{RAMEN} & \multirow{4}{*}{1024} & \multirow{4}{*}{3} & \multirow{2}{*}{512}
& 96 & 27.64 & 32.79 & 808.81 & 70.16\\
& & & & 48 & 30.84 & 35.78 & 3170.16 & 197.36\\
\cmidrule(lr){4-9}
& & & \multirow{2}{*}{256} & 24 & 31.74 & 36.49 & 12622.58 & 658.30\\
& & & & 12 & \textbf{32.19} & 39.85 & 50404.06 & 3658.17\\
\midrule
DOFA & & & &  & & 39.29 & 3760.22 & 230.98\\
TerraMind-B & & & & &  & 37.87 & 3873.00 & 205.79\\
TerraMind-L & & & & &  & 37.89 & 10591.81 & 625.38\\
\bottomrule
\end{tabular}
\end{adjustbox}
\label{tab:dyn_ltae}
\end{table*}

\begin{table*}[!t]
\centering
\small
\caption{\textbf{CropTypeMapping- South Sudan with RAMEN temporal encoder performances.} We report validation and test mIoU, along with GFLOPs and inference time per tile for RAMEN evaluated at different $\text{GSD}_{target}$.} 
\begin{adjustbox}{width=1\textwidth}
\begin{tabular}{lccccc|cccc}
\toprule
Model & Modality & Image size & $\text{GSD}_m$ & Input size & $\text{GSD}_{target}$ 
& val mIoU & test mIoU & GFLOPS & Inference time/tile (ms)\\
\midrule
\multirow{8}{*}{RAMEN} & \multirow{4}{*}{S2} & \multirow{4}{*}{64} & \multirow{4}{*}{10} & \multirow{4}{*}{64} 
& 80 & 46.81 & 50.69 & 21.65 & 8.68\\
& & & & & 60 & 45.89 & 33.03 & 33.65 & 8.64\\
& & & & & 40 & 54.13 & 46.21 & 85.66 & 9.78\\
& & & & & 20 & \textbf{57.20} & 53.01 & 341.68 & 17.03\\
\cmidrule(lr){2-10}
& \multirow{4}{*}{S2+S1} & \multirow{4}{*}{64} & \multirow{4}{*}{10} & \multirow{4}{*}{64}
& 80 & 50.48 & 45.09 & 31.86 & 12.37\\
& & & & & 60 & 47.80 & 47.93 & 49.57 & 13.00\\
& & & & & 40 & 47.25 & 51.92 & 126.31 & 14.98\\
& & & & & 20 & 51.00 & 44.82 & 504.08 & 28.77\\
\midrule
DOFA & & & & & & & 51.33 & 153.49 & 27.71\\
TerraMind-B & & & & & & & 55.80 & 268.12 & 24.08\\
TerraMind-L & & & & & & & 55.04 & 813.65 & 66.19\\
\bottomrule
\end{tabular}
\end{adjustbox}
\label{tab:CTM_no_ltae}
\end{table*}

\begin{table*}[!t]
\centering
\small
\caption{\textbf{CropTypeMapping - South Sudan with late LTAE fusion performances.} We report validation and test mIoU, along with GFLOPs and inference time per tile for RAMEN evaluated at different $\text{GSD}_{target}$.} 
\begin{adjustbox}{width=1\textwidth}
\begin{tabular}{lccccc|cccc}
\toprule
Model & Modality & Img size & $\text{GSD}_m$ & Input size & $\text{GSD}_{target}$ 
& val mIoU & test mIoU & GFLOPs & Inference time/tile (ms)\\
\midrule
\multirow{8}{*}{RAMEN} & \multirow{4}{*}{S2} & \multirow{4}{*}{64} & \multirow{4}{*}{10} & \multirow{4}{*}{64} 
& 80 & 47.26 & 49.57 & 50.05 & 24.58\\
& & & & & 60 & 46.71 & 39.92 & 77.73 & 26.37\\
& & & & & 40 & 53.85 & 46.31 & 197.64 & 31.59\\
& & & & & 20 & 56.95 & 53.85 & 787.96 & 61.81\\
\cmidrule(lr){2-10}
& \multirow{4}{*}{S2+S1} & \multirow{4}{*}{64} & \multirow{4}{*}{10} & \multirow{4}{*}{64}
& 80 & 58.19 & 52.83 & 90.26 & 31.83\\
& & & & & 60 & 59.92 & 50.53 & 140.50 & 34.93\\
& & & & & 40 & \textbf{62.22} & 53.27 & 358.17 & 47.22\\
& & & & & 20 & 57.35 & 41.68 & 1429.94 & 109.66\\
\midrule
DOFA & & & & & & & 51.33 & 153.49 & 27.71\\
TerraMind-B & & & & & & & 55.80 & 268.12 & 24.08\\
TerraMind-L & & & & & & & 55.04 & 813.65 & 66.19\\
\bottomrule
\end{tabular}
\end{adjustbox}
\label{tab:CTM_ltae}
\end{table*}

\begin{table*}[!t]
\centering
\small
\caption{\textbf{SpaceNet 7 performances.} We report validation and test mIoU, along with GFLOPs and inference time per tile for RAMEN evaluated at different $\text{GSD}_{target}$.} 
\begin{adjustbox}{width=1\textwidth}
\begin{tabular}{lcccc|cccc}
\toprule
Model & Img size & $\text{GSD}_m$ & Input size & $\text{GSD}_{target}$ & val mIoU & test mIoU & GFLOPs & Inference time/tile (ms)\\
\midrule
\multirow{5}{*}{RAMEN} & \multirow{5}{*}{256} & \multirow{5}{*}{4} & \multirow{2}{*}{256} & 64 & 50.63 & 52.80 & 80.60 & 7.40\\
& & & & 32 & 53.93 & 55.46 & 321.61 & 13.93\\
\cmidrule(lr){4-9}
& & & 128 & 16 & 58.69 & 59.62 & 1285.97 & 41.42\\
\cmidrule(lr){4-9}
& & & 64 & 8 & \textbf{59.48} & 60.31 & 5143.45 & 152.53 \\
\midrule
DOFA & & & & & & 61.84 & 245.67 & 13.75 \\
TerraMind-B & & & & & & 60.61 & 248.94 & 12.91 \\
TerraMind-L & & & & & & 59.98 & 435.47 & 24.42\\
\bottomrule
\end{tabular}
\end{adjustbox}
\label{tab:SpaceNet7}  
\end{table*}

\begin{table*}[!t]
\centering
\small
\caption{\textbf{AI4SmallFarms performances.} We report \textbf{only test} mIoU (no validation set provided), along with GFLOPs and inference time per tile for RAMEN evaluated at different $\text{GSD}_{target}$.} 
\begin{adjustbox}{width=1\textwidth}
\begin{tabular}{lcccc|ccc}
\toprule
Model & Img size & $\text{GSD}_m$ & Input size & $\text{GSD}_{target}$ & test mIoU & GFLOPs & Inference time/tile (ms)\\
\midrule
\multirow{5}{*}{RAMEN} & \multirow{5}{*}{496} & \multirow{5}{*}{10} & 496 & 80 & 26.98 & 1207.16 & 48.39\\
\cmidrule(lr){4-8}
& & & 248 & 40 & 30.00 & 4826.39 & 168.77\\
\cmidrule(lr){4-8}
& & & 124 & 20 & 35.35 & 19303.31 & 658.78\\
\cmidrule(lr){4-8}
& & & 64 & 10 & \textbf{38.78} & 77210.97 & 2617.79\\
\midrule
DOFA & & & & & 27.07 & 553.33 & 29.71 \\
TerraMind-B & & & & & 28.12 & 560.11 & 27.37 \\
TerraMind-L & & & & & 27.47 & 980.05 & 27.47\\
\bottomrule
\end{tabular}
\end{adjustbox}
\label{tab:ai4smallfarms}  
\end{table*}

\section{Qualitative results}
\label{sup:qualitative}

We present in Fig.~\ref{fig:qualitative_results_all} a set of qualitative examples of the predicted segmentation maps at various $\text{GSD}_{target}$ values across four downstream tasks. These results showcase how increasing spatial resolution enables RAMEN to produce fine-grained representations on pixel-level segmentation tasks (e.g. AI4SmallFarms).

\begin{figure*}[!t]
\centering
\begin{subfigure}{0.85\textwidth}
    \centering
    \includegraphics[width=\textwidth]{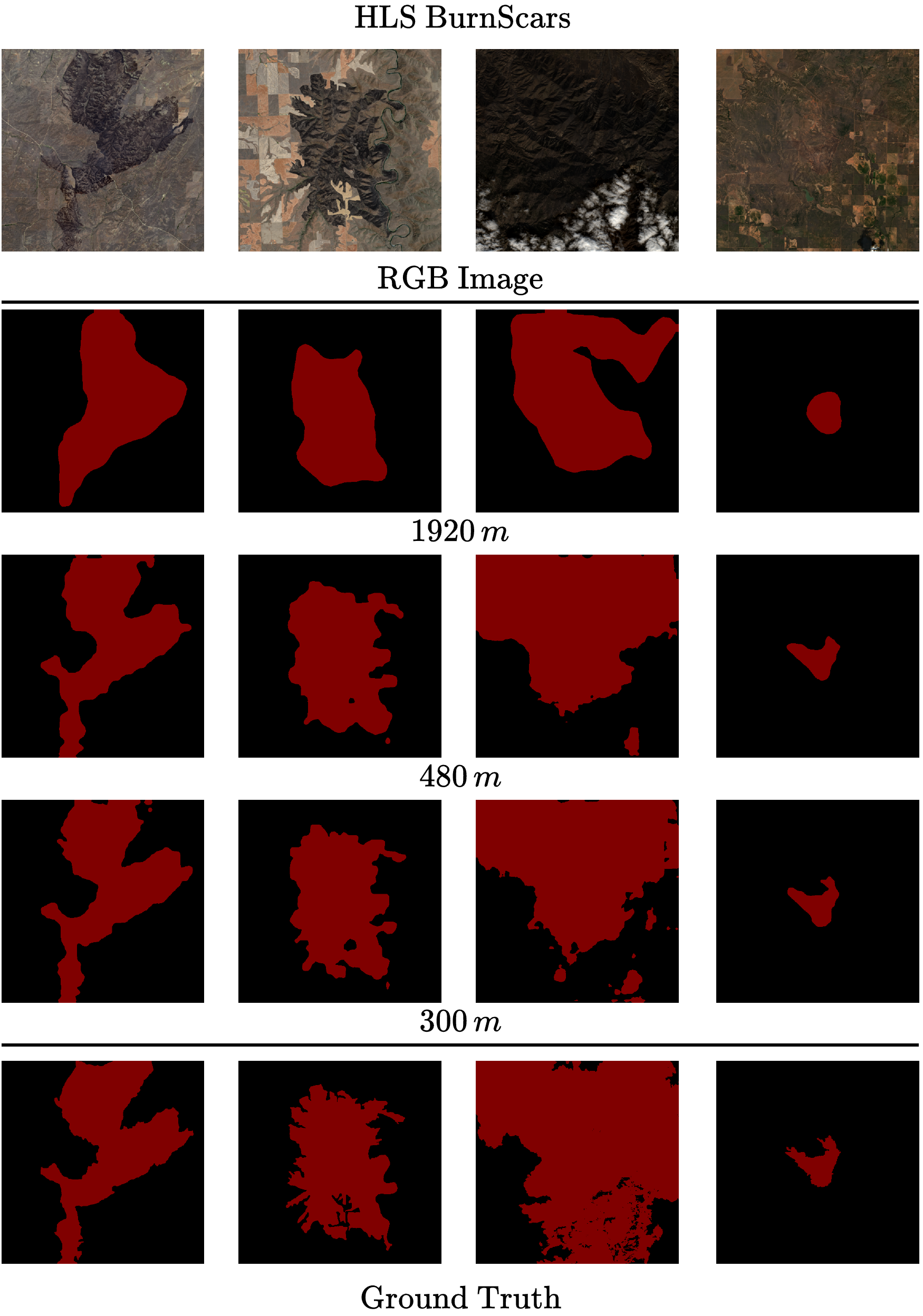}
    \caption{HLS BurnScars}
    \label{fig:qualitative_results_burnsr}
\end{subfigure}

\caption{\textbf{Illustrative prediction results across datasets.} Segmentation maps produced at various coarse to fine $\text{GSD}_{target}$.}
\label{fig:qualitative_results_all}
\end{figure*}

\clearpage

\begin{figure*}[!t]
\ContinuedFloat
\centering
\begin{subfigure}{0.85\textwidth}
    \centering
    \includegraphics[width=\textwidth]{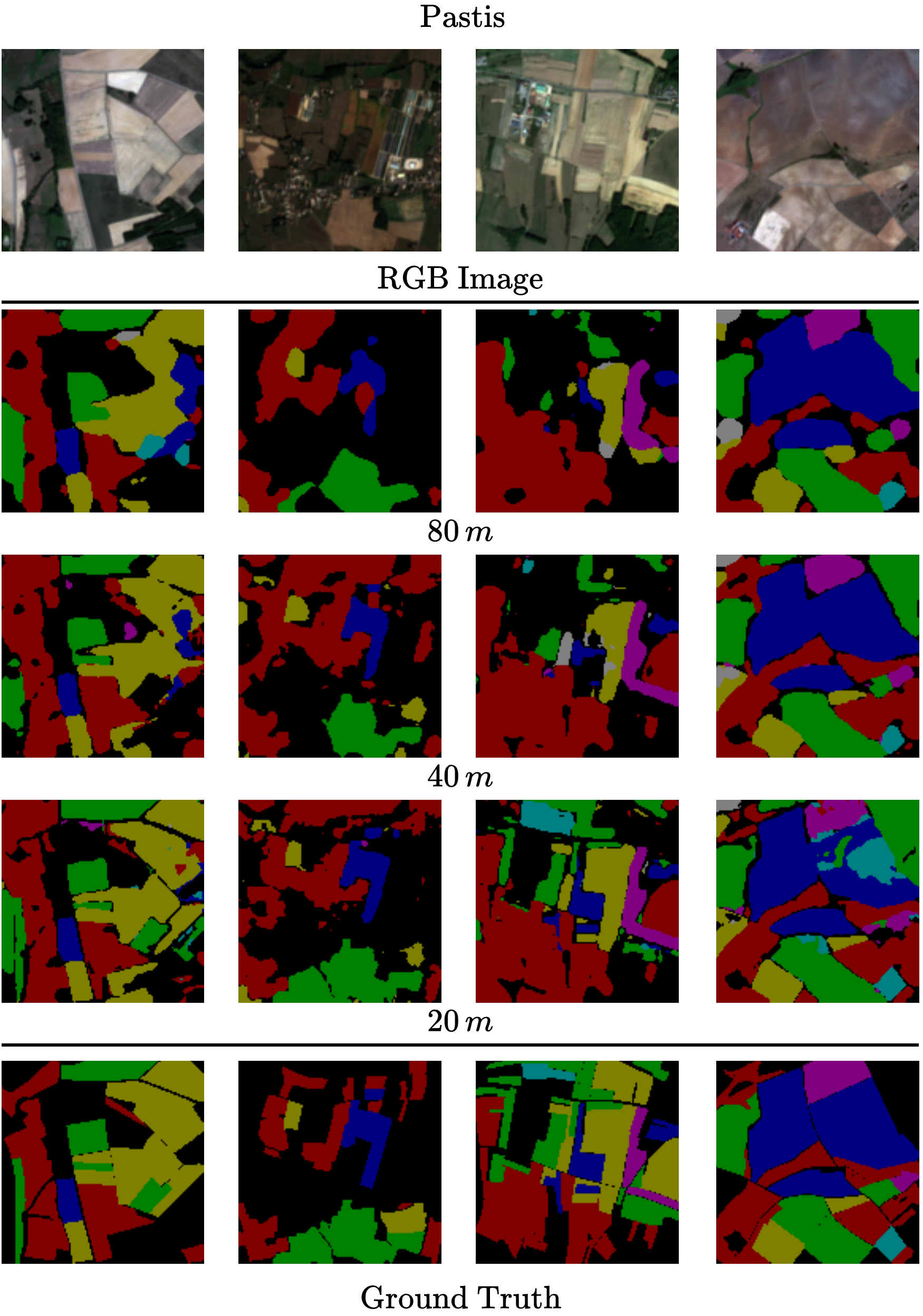}
    \caption{Pastis}
    \label{fig:qualitative_results_pastis}
\end{subfigure}
\end{figure*}

\clearpage

\begin{figure*}[!t]
\ContinuedFloat
\centering
\begin{subfigure}{0.85\textwidth}
    \centering
    \includegraphics[width=\textwidth]{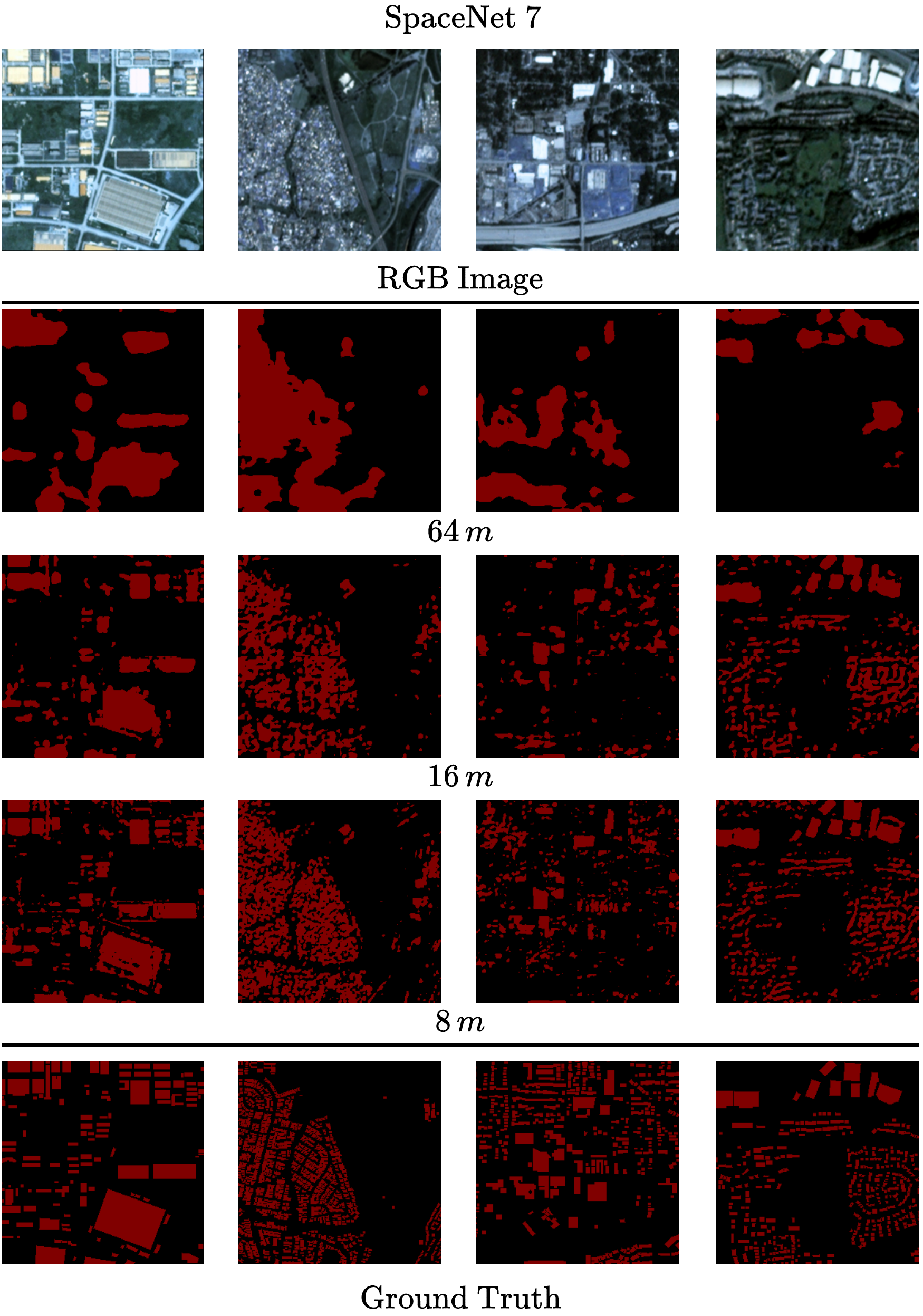}
    \caption{SpaceNet 7}
    \label{fig:qualitative_results_sn7}
\end{subfigure}
\end{figure*}

\clearpage

\begin{figure*}[!t]
\ContinuedFloat
\centering
\begin{subfigure}{0.85\textwidth}
    \centering
    \includegraphics[width=\textwidth]{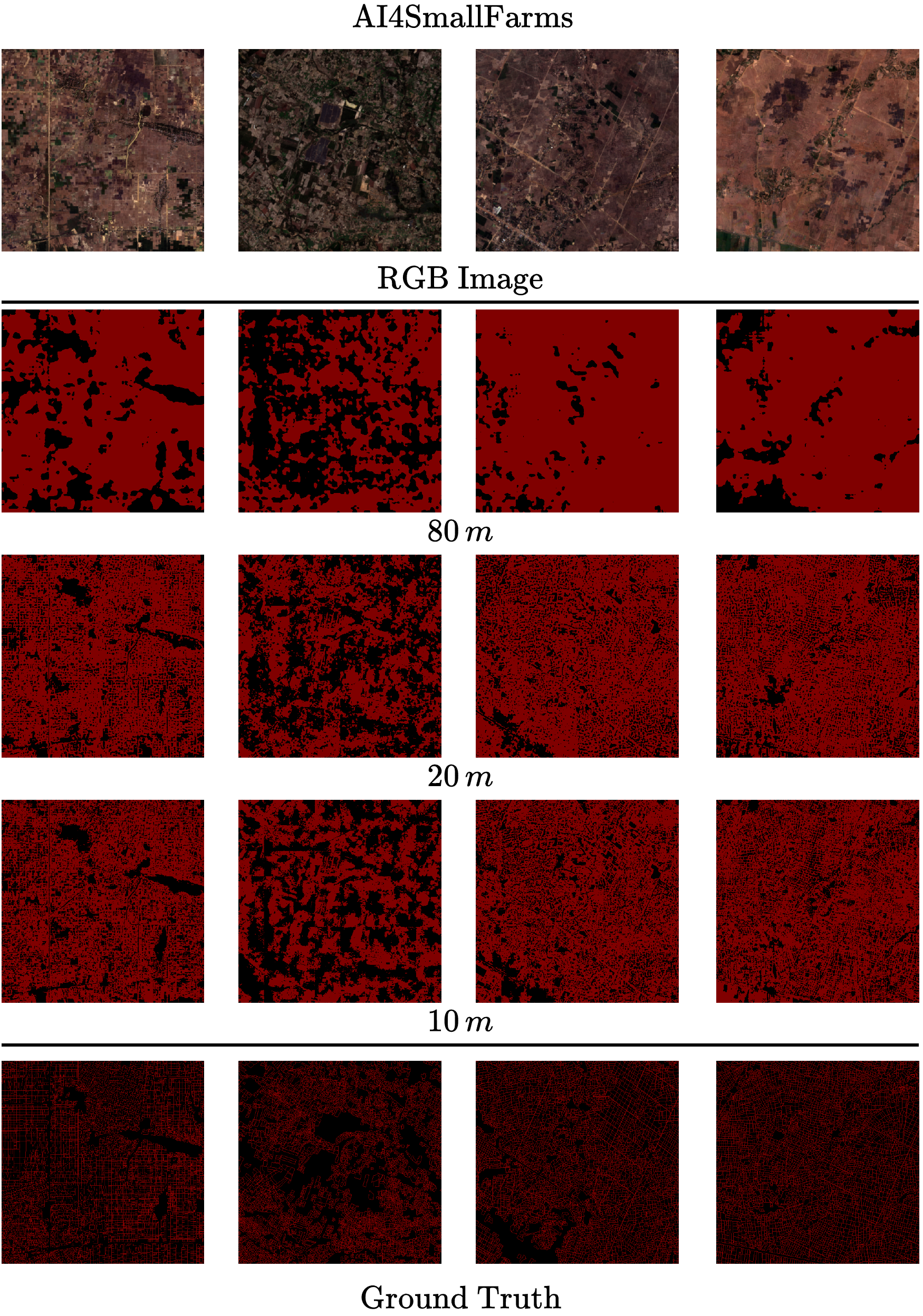}
    \caption{AI4SmallFarms}
    \label{fig:qualitative_results_ai4small}
\end{subfigure}
\end{figure*}

\end{document}